\pdfoutput=1

\documentclass[10pt,journal,compsoc]{IEEEtran}

\usepackage[nocompress]{cite}
\usepackage[pdftex]{graphicx}
\usepackage{amsmath, amssymb}
\usepackage{xcolor}
\usepackage{multirow}
\usepackage[colorlinks = true, citecolor=blue]{hyperref}

\usepackage{paralist}

\newcommand{\ytyy}[1]{\textcolor{black}{#1}}
\newcommand{\yty}[1]{\textcolor{black}{#1}}
\newcommand{\tyy}[1]{\textcolor{black}{#1}}
\newcommand{\tyyy}[1]{\textcolor{black}{#1}}
\newcommand{\first}[1]{\textcolor{red}{#1}}
\newcommand{\second}[1]{\textcolor{green}{#1}}
\newcommand{\third}[1]{\textcolor{blue}{#1}}
\newcommand{\abc}[1]{\textcolor{black}{#1}}
\newcommand{\abcn}[1]{\textcolor{black}{#1}}

\newcommand{\rt}[1]{\underline{#1}}
\DeclareMathOperator*{\argmax}{argmax}
\def\baselinestretch{0.975}

\begin{document}

\title{Visual Tracking via Dynamic Memory Networks}

\author{Tianyu Yang and Antoni B. Chan
\IEEEcompsocitemizethanks{\IEEEcompsocthanksitem The authors are with the Department of Computer Science, City University of Hong Kong.\protect\\
E-mail: tianyyang8-c@my.cityu.edu.hk, abchan@cityu.edu.hk}}

\markboth{Journal of \LaTeX\ Class Files,~Vol.~14, No.~8, August~2015}%
{Shell \MakeLowercase{\textit{et al.}}: Bare Demo of IEEEtran.cls for Computer Society Journals}

\IEEEtitleabstractindextext{%
\begin{abstract}
Template-matching methods for visual tracking have gained
popularity recently due to their \abc{good} 
performance and fast speed.
However, they lack effective ways to adapt to changes in the target object's appearance, making their tracking accuracy still far from state-of-the-art. In this paper, we propose a dynamic memory network to adapt the template to the target's appearance variations during tracking. \yty{The reading and writing process of the external memory is controlled by an LSTM network with the search feature map as input. A spatial attention mechanism is applied to concentrate the LSTM input on the potential target as the location of the target is at first unknown}. 
To prevent aggressive model adaptivity, we apply gated residual template learning to control the amount of retrieved memory that is used to combine with the initial template. \yty{In order to alleviate the drift problem, we also design a \abc{``negative'' memory unit that stores templates for distractors, which are used to cancel out wrong responses from the object template.}
%
To further boost the tracking performance,  an auxiliary classification loss is added after the feature extractor part.} Unlike tracking-by-detection methods where the object's information is maintained by the weight parameters of neural networks, which requires expensive online fine-tuning to be adaptable, our tracker runs completely feed-forward and adapts to the target's appearance changes by updating the external memory. Moreover, the capacity of our model is not determined by the network size as with other trackers --  the capacity can be easily enlarged as the memory requirements of a task increase, which is favorable for memorizing long-term object information. Extensive experiments on the OTB and VOT datasets demonstrate that our trackers perform favorably against state-of-the-art tracking methods
while retaining real-time speed.
\end{abstract}

}

\maketitle

\IEEEdisplaynontitleabstractindextext

%
\IEEEpeerreviewmaketitle

\IEEEraisesectionheading{\section{Introduction}\label{sec:introduction}}

\IEEEPARstart{R}{ecent} \yty{years have witnessed the great success of convolution neural networks (CNNs) applied to image recognition \cite{Krizhevsky2012, Szegedy2015, He2016}, object detection \cite{Girshick2014, Girshick2015, Ren2015} and semantic segmentation \cite{Long2015, Noh2016, Li2017}. The visual tracking community also \abc{sees} an increasing number of trackers \cite{Song2017, Nam2016, Wang2015, Bertinetto2016, Guo2017}  adopting deep learning models to boost their performance.} Among them are two dominant tracking strategies. One is the {\em tracking-by-detection} scheme that online trains an object appearance classifier \cite{Song2017, Nam2016} to distinguish the target from the background. The model is first learned using the initial frame, and then fine-tuned using the training samples generated in the subsequent frames based on the newly predicted bounding box. The other scheme is {\em template matching}, which adopts either the target patch in the first frame \cite{Bertinetto2016, Tao2016} or the previous frame \cite{Held2016} to construct the matching model. To handle changes in the target appearance, 
the template built in the first frame may be interpolated by the recently generated object template with a small learning rate \cite{Valmadre2017}.  

The main difference between these two strategies is that tracking-by-detection maintains the target's appearance information in the weights of the deep neural network, thus requiring online fine-tuning with stochastic gradient descent (SGD) to make the model adaptable,
while in contrast, template matching stores the target's appearance in the object template, which is generated by feed forward computations.  Due to the computationally expensive model updating required in tracking-by-detection, the speed of such methods are usually slow, e.g.\ 
\cite{Song2017, Nam2016, Nam2016-1} run at about 1 fps,
although they do achieve state-of-the-art tracking accuracy. 
%
Template matching methods, however, are fast 
because there is no need to update the parameters of the neural networks. Recently, several trackers \cite{Bertinetto2016, Guo2017, Yang2017, He2018, Wang2018} adopt fully convolutional Siamese networks as the matching model, which demonstrate promising results and real-time speed.  However, there is still a large performance gap between template-matching models and tracking-by-detection, due to the lack of an effective method for adapting to appearance variations online.

\begin{figure*}[t]
	\begin{center}
		\includegraphics[width=0.95\linewidth]{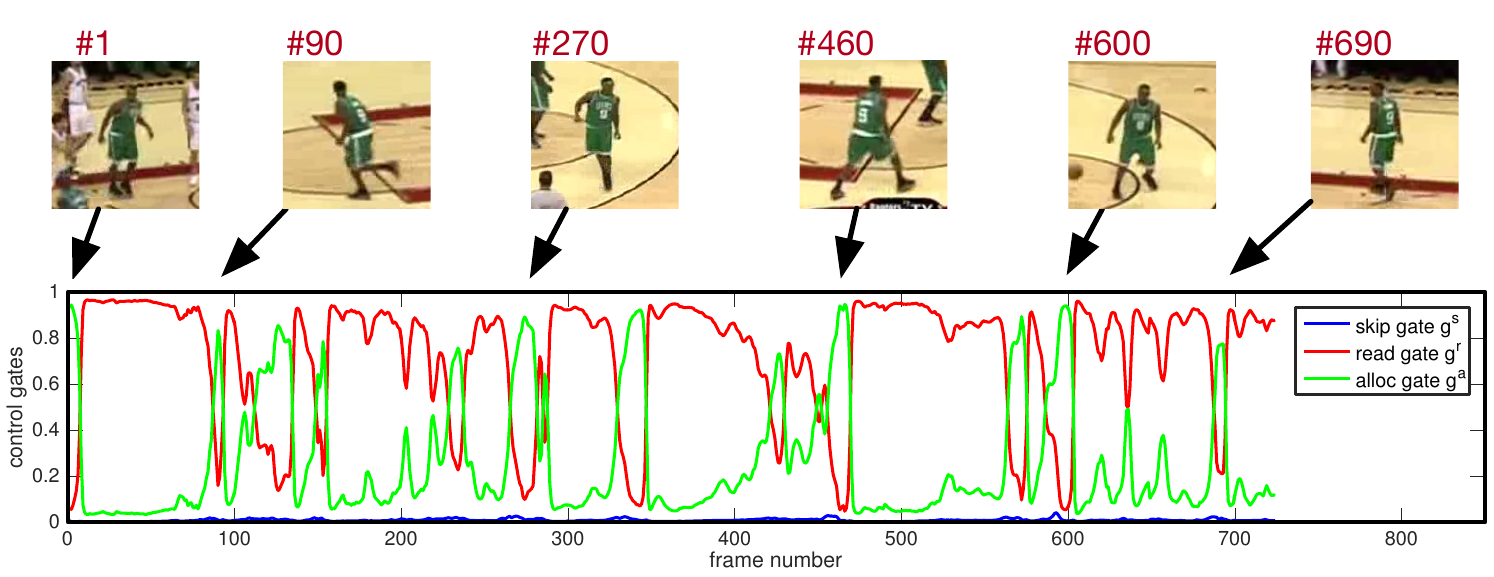}
	\end{center}
	\vspace{-6mm}
	\caption{Example of template updating on the Basketball video: the control gate signals change along with the appearance variations. When there are large appearance changes, the allocation gate approaches to 1, which means a new memory slot is overwritten. 
	\abc{When there are only small appearance variations in the object template, then the read gate is close to 1, which indicates that the most recently read memory slot will be updated.}
See Section \ref{memwrite} for detailed explanations.
	}
	\vspace{-5mm}
	\label{fig:1}
\end{figure*}

In this paper, we propose a dynamic memory network, where the target information is stored and recalled from  external memory,  to maintain the variations of object appearance for template-matching (See an example in Figure \ref{fig:1}).
Unlike tracking-by-detection  where the target's information is stored in the weights of neural networks and \yty{therefore the capacity of the model is fixed by the number of parameters}, the model capacity of our memory networks can be easily enlarged by increasing the size of external memory, which is useful for memorizing long-term appearance variations. 
Since aggressive template updating is prone to overfit recent frames and the initial template is the most reliable one,
we use the initial template as a conservative reference of the object and a residual template, 
obtained from retrieved memory, to adapt to the appearance variations.
During tracking, the residual template is 
gated channel-wise and 
combined with the initial template to form the \yty{positive matching template.}
The channel-wise gating of the residual template controls how much each channel of the retrieved template should be added to the initial template, which can be interpreted as a feature/part selector for adapting the template. 
\abc{Besides the positive template, a second ``negative'' memory unit stores templates of potential distractor objects.  The negative template is used to cancel out non-discriminative channels (corresponding to object parts) in the positive template, yielding the final template, which is convolved with the search image features to get the response map.}
%
\yty{The reading and writing process of the positive and negative memories, as well as the channel-wise gate vector for the residual template, is controlled by an LSTM (Long Short-Term Memory) whose input is based on the search feature map.}
As the target position is at first unknown in the search image, we adopt an attention mechanism to locate the object roughly in the search image, thus leading to a soft representation of the target for the input to the LSTM controller. This helps to retrieve the most-related template in the memory. \yty{In addition, we further improve the tracking performance by adding an auxiliary classification loss at the end of the CNN feature extractor, which is aimed at improving the tracker's robustness to appearance variations. \abc{The tracking  and classification losses serve complementary roles} -- learning features through similarity matching facilitates their ability of precise localization, while training features on the auxiliary classification problem 
provides semantic information for tracking robustness.}
The whole framework is differentiable and therefore can be trained end-to-end with SGD. In summary, the contributions of our work are:
\begin{compactitem}
	\item We design a dynamic memory network for visual tracking. An external memory block, which is controlled by an LSTM with attention mechanism, allows adaptation to appearance variations. 
	\item We propose gated residual template learning to generate the final matching template, which effectively controls the amount of appearance variations in retrieved memory that is added to each channel of the initial matching template.
	This prevents excessive model updating, while retaining the  conservative information of the target.
	\yty{\item We propose a negative template memory for storing and retrieving distractor templates, which are used to cancel the response peaks due to distractor objects, thus alleviating 
	 drift problems caused by distractors.}
	\yty{\item We add an auxiliary classification branch 
	after the feature extraction block, \abc{which trains the features to also contain semantic information.  This increases the robustness of the features to variations in object appearances, 
	and boosts the tracking performance.}}
	\item We extensively evaluate our algorithm on large scale datasets OTB and VOT. Our trackers perform favorably against state-of-the-art tracking methods while possessing real-time speed.
\end{compactitem}

\tyy{The remainder of the paper is organized as follows. In Section 2, we briefly review related work. 
In Section 3, we describe our proposed tracking methods, and in Section 4 we present  implementation details. We perform extensive experiments on OTB and VOT datasets in Section 5.}

\section{Related Work}

\abc{In this section, we review related work on tracking-by-detection, tracking by template-matching, memory networks and multi-task learning.}
\abc{A preliminary version of our work appears in ECCV 2018 \cite{Yang2018}. This paper contains additional improvements in both methodology and experiments, including: 
1) \tyy{we propose a negative memory unit that stores distractor templates to cancel out wrong responses from the object template; 
2) we design an auxiliary classification loss to facilitate the tracker's robustness to 
appearance changes; 
3) we conduct comprehensive experiments on the VOT datasets, including VOT-2015, VOT-2016 and VOT-2017.}}

\subsection{Tracking by Detection}
\yty{Tracking-by-detection treats object tracking as a detection problem within \ytyy{an} ROI image, where an online learned classifier is used to distinguish the target from the background.
The difficulty 
\ytyy{of} updating the classifier to adapt to appearance variations is that the bounding box predicted on each frame may not be accurate, which produces degraded training samples and thus gradually causes the tracker to drift. 
Numerous algorithms have been designed to mitigate the sample ambiguity caused by inaccurate predicted bounding boxes. \cite{Grabner2008} formulates the online model learning process in a semi-supervised fashion by combining a given prior and the trained classifier. \cite{Babenko2011} proposes a multiple instance learning scheme to solve the problem of inaccurate examples 
for online training. Instead of only focusing on facilitating the training process of the tracker, 
\cite{Kalal2012} decomposes the tracking task into three parts---tracking, learning and detection, where a optical flow tracker is used for frame-to-frame tracking and an online trained detector is adopted to re-detect the target when drifting occurs. 
}

\yty{With the widespread use of CNNs in the computer vision community, many methods \cite{li2018deep} have applied CNNs as the classifier to localize the target.
\cite{Wang2015} uses two fully convolutional neural networks to estimate the target's bounding box, including a GNet that captures category information and an SNet that classifies the target from the background. \cite{Nam2016} presents a multi-domain learning framework to learn the shared representation of objects from different sequences. Motived by Dropout \cite{Srivastava2014}, BranchOut \cite{Han2017} adopts multiple branches of fully connected layers, from which  a random subset are selected for training, which regularizes the neural networks to avoid overfitting. Unlike these tracking-by-detection algorithms, which need costly stochastic gradient decent (SGD) updating, our method runs completely feed-forward and adapts to the object's appearance variations through a memory writing process, \abc{thus achieving real-time performance.}}

\subsection{Tracking by Template-Matching} Matching-based methods have recently gained popularity due to their fast speed and 
\tyy{promising} performance. 
The most notable is the fully convolutional Siamese network (SiamFC) \cite{Bertinetto2016}. Although it only uses the first frame as the template, SiamFC achieves competitive results and fast speed. The key deficiency of SiamFC is that it lacks an effective model for online updating. 
To address this, \cite{Valmadre2017} updates the model using linear interpolation of new templates with a small learning rate, but  only 
sees modest improvements in accuracy.
RFL (Recurrent Filter Learning)  \cite{Yang2017} adopts a convolutional LSTM for model updating, where the forget and input gates control the linear combination of the historical target information (\emph{i.e.}, memory states of the LSTM) and the object's current template automatically. Guo \emph{et al.} \cite{Guo2017} propose a dynamic Siamese network with two general transformations for target appearance variation and background suppression. \ytyy{He \emph{et. al.} \cite{He2018} design two branches of Siamese networks with a channel-wise attention mechanism aiming to improve the robustness and discrimination ability of the matching network.}

To further improve the speed of SiamFC, \cite{Huang2017} 
reduces the feature computation cost for easy frames, by using deep reinforcement learning to train policies for early stopping the feed-forward calculations of the CNN when the response confidence is high enough.
%
SINT \cite{Tao2016} also uses Siamese networks for visual tracking and has higher accuracy, but runs much slower than SiamFC (2 fps vs 86 fps) due to the use of a deeper CNN (VGG16) for feature extraction, and optical flow for its candidate sampling strategy. \tyyy{\cite{chi2017dual} proposes a dual deep network by exploiting hierarchical features of CNN layers for object tracking.}  Unlike other template-matching models that use sliding windows or random sampling to generate candidate image patches for testing, GOTURN \cite{Held2016} directly regresses the coordinates of the target's bounding box by comparing the previous and current image patches.  \ytyy{Despite its fast speed and advantage on handling scale and aspect ratio changes}, its tracking accuracy is much lower than other state-of-the-art trackers. 

Different from existing matching-based trackers where the capacity to adapt is limited by the neural network size, we use  SiamFC 
as the baseline feature extractor and 
add
an addressable memory, whose memory size is independent of the neural networks and thus can be easily enlarged as memory requirements of a tracking task increase.

\begin{figure*}[t]
	\begin{center}
		\includegraphics[width=0.95\linewidth]{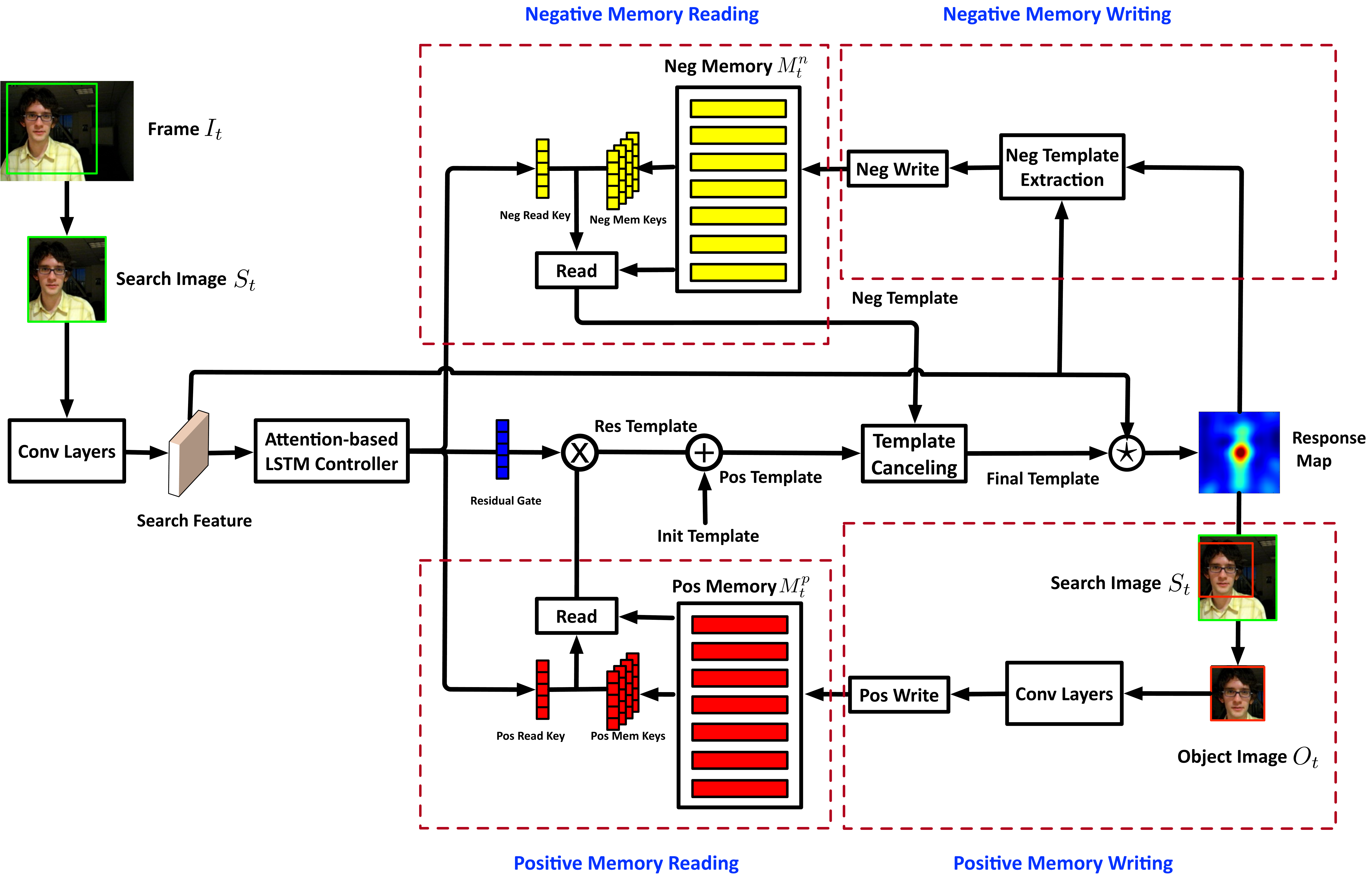}
	\end{center}
	\vspace{-5mm}
	\caption{The pipeline of our tracking algorithm. The green rectangle is the candidate region for target searching. The \textit{Feature Extraction} blocks for the object image and search image share the same architecture and parameters. An attentional LSTM extracts the target's information on the search feature map, which guides the memory reading process to retrieve a matching template.  The residual  template is combined with the initial template, to obtain \yty{a positive template. 
			\abc{A negative template is read from the negative memory and combined with the positive template to cancel responses from distractor objects.
				The final template is convolved with the search feature map to obtain the response map.}
			The newly predicted bounding box is then used to crop the object's \abc{feature map} 
			for writing to the positive memory.  A negative template is extracted from the search feature map using the response score and written to negative memory.}
	}
	\vspace{-3mm}
	\label{fig:2}
\end{figure*}

\subsection{Memory Networks} The recent use of convolutional LSTM for visual tracking \cite{Yang2017} shows that memory states 
are useful for object template management over long timescales. Memory networks are typically used to solve simple logical reasoning problem in natural language processing (NLP), e.g., question answering and sentiment analysis. The pioneering works include NTM (Neural Turing Machine) \cite{Graves2014} and MemNN (Memory Neural Networks) \cite{Weston2015}. They both propose an addressable external memory with reading and writing mechanisms -- NTM focuses on problems of sorting, copying and recall, while MemNN aims at language and reasoning tasks. MemN2N 
\cite{Sukhbaatar2015} further improves MemNN by removing the supervision of supporting facts, which makes it trainable in an end-to-end fashion. Based on 
NTM, 
\cite{Graves2016} proposes DNC (Differentiable Neural Computer), which uses a different access mechanism to alleviate the memory overlap and interference problems.
Recently, NTM is also applied to one-shot learning \cite{Santoro2016} by redesigning the method for reading and writing memory, and has shown promising results at 
encoding and retrieving new information quickly. 
\ytyy{\cite{Liu2017} also proposes a memory-augmented tracking algorithm, which obtains limited performance and lower speed (5 fps) due to two reasons.
First, in contrast to our method, they performs dimensionality reduction of the object template (from 20x20x256 to 256) when storing it into memory, resulting in loss of spatial information for template matching. Second, they extract multiple patches centered on different positions of the search image to retrieve the proper memory, which is not efficient compared with our attention scheme.}

Our proposed memory model differs from the aforementioned memory networks in the following aspects. First, for the question answering problem, the input of each time step is a sentence,
\emph{i.e.}, a sequence of feature vectors (each word corresponds to one vector) that needs an embedding layer (usually RNN) to obtain an internal state. In contrast, for object tracking, the input is a search image that needs a feature extraction process (usually CNN) to get a more abstract representation. Furthermore, for object tracking, the target's position in the search image patch is unknown, and here we propose an attention mechanism to highlight the target's information when generating the read key for memory retrieval. 
Second, the dimension of feature vectors stored in memory for NLP is relatively small (50 in MemN2N vs.~6$\times$6$\times$256=9216 in our case). 
Directly using the original template for address calculation is time-consuming.  Therefore we apply an average pooling on the feature map to generate a template key for addressing, which is efficient and effective experimentally. 
Furthermore, we apply channel-wise gated residual template learning for model updating, and redesign the memory writing operation to be more suitable for visual tracking.
	

\ytyy{\subsection{Multi-task learning}
Multi-task learning has been successfully used in many applications of machine learning, ranging from natural language processing \cite{collobert2008unified} and speech recognition \cite{deng2013new}  to computer vision  \cite{girshick2015fast}. \cite{caruana1997multitask}  estimates the street direction in an autonomous driving car by predicting various characteristics of the road, which serves as an auxiliary task.  \cite{zhang2012convex}  introduces auxiliary tasks of estimating head pose and facial attributes to boost the performance of facial landmark detection, while \cite{li2015heterogeneous} boosted the performance of a human pose estimation network by adding human joint detectors as auxiliary tasks. Recent works combining object detection and semantic segmentation \cite{yao2012describing, He2017}, as well as image depth estimation and semantic segmentation \cite{eigen2015predicting, kendall2018multi}, also demonstrate the effectiveness of multi-task learning on improving the generalization ability of neural networks. Observing that the CNN learned for object similarity matching lacks the generalization ability of invariance to appearance variations, we propose to add an auxiliary task,  object classification, to regularize the CNN so that it learns object semantics.}

\section{Dynamic Memory Networks for Tracking}

In this section, we propose a dynamic memory network with reading and writing mechanisms for visual tracking. 
The whole framework is shown in Figure \ref{fig:2}.
Given the search image, first features are extracted with a CNN.
The image features are input into an attentional LSTM, which controls memory reading and writing. 
A residual template is read from the \yty{positive memory} and combined with the initial template learned from the first frame, forming the \yty{ positive template. Then a negative template is retrieved from the negative memory to cancel parts of the positive template through a channel-wise gate, forming the final template.} The final template is convolved with the search image features to obtain the response map, and the target bounding box is predicted.
The new target's template is cropped using the predicted bounding box, features are extracted and then written into \yty{positive }memory for model updating. \yty{The negative template is extracted on the search feature map based on the response map. Responses whose corresponding score is greater than a threshold and whose distance are far from the target's center are considered as negative (distractor) templates for negative memory writing.}

\subsection{Feature Extraction}

Given an input image $I_t$ at time $t$, we first crop the frame into a search image patch $S_t$ with a rectangle that is computed from the previous predicted bounding box \yty{as in \cite{Bertinetto2016}}.
Then it is encoded into a high level representation $f(S_t)$, which is a spatial feature map, via a fully convolutional neural networks (FCNN).  In this work we use the FCNN structure from SiamFC \cite{Bertinetto2016}. 
After getting the predicted bounding box, we use the same feature extractor to compute the new object template for \yty{positive memory} writing.

\begin{figure}[t]
	\begin{center}
		\includegraphics[width=\linewidth]{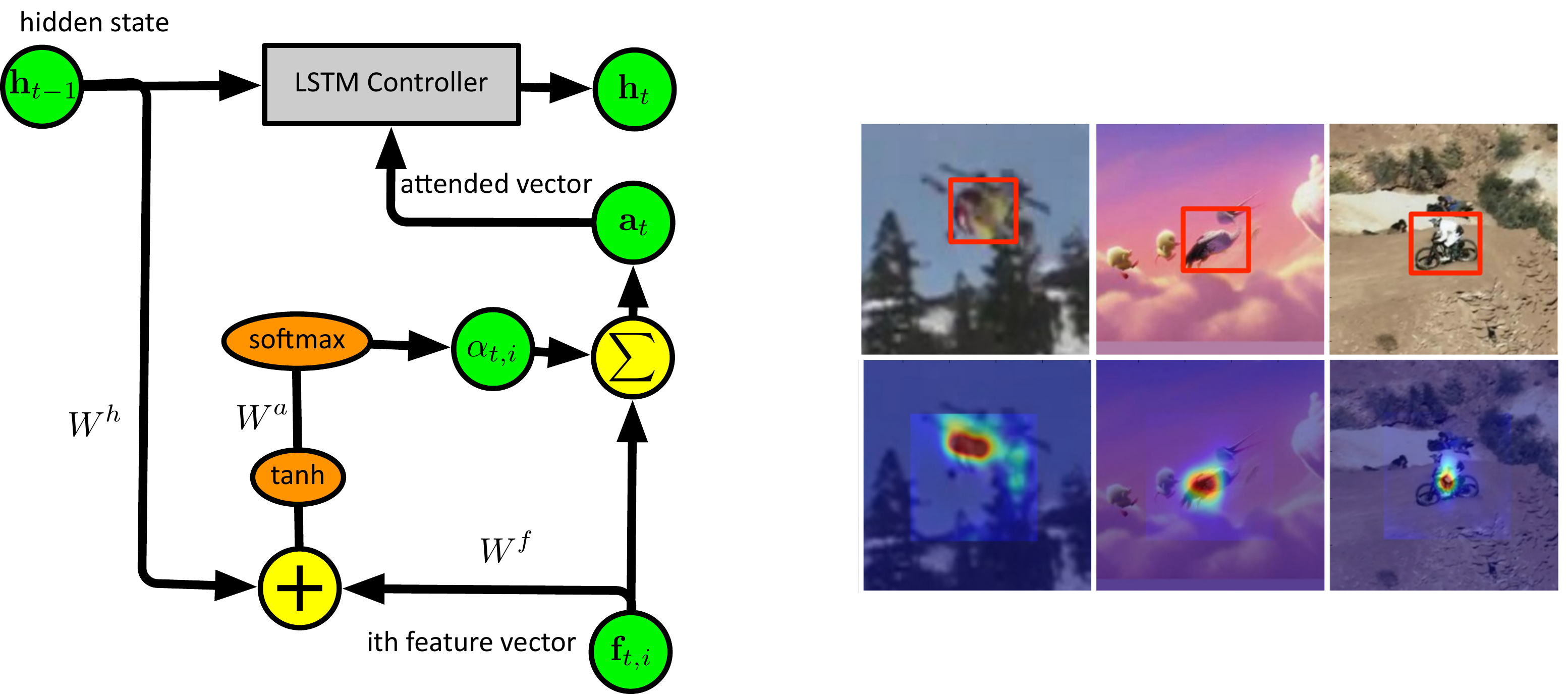}
	\end{center}
	\vspace{-5mm}
	\caption{\textbf{Left}: Diagram of attention  network.  \textbf{Right}: Visualization of attentional weights map: for each pair, \abc{(top)} search images and ground-truth target box, and \abc{(bottom)} attention maps over search image. For visualization, the attention maps are resized using bicubic interpolation to match the size of the original image.}
	\label{fig:3}
	\vspace{-4mm}
\end{figure}


\subsection{Attention Scheme}\label{attention}

Since the object information in the search image is needed to retrieve the related template for matching, but the object location is unknown at first, we apply an attention mechanism to make the input to the LSTM concentrate more on the target.
We define $F_{t,i} \in \mathbb{R}^{n \times n \times c}$ as the $i$-th $\mathit{n\times n\times c}$ square patch on $F_t=f(S_t)$ in a sliding window fashion.\footnote{We use $6\times6\times256$, which is the same size of the matching template.}
Each square patch covers a certain part of the search image. An attention-based weighted sum of these square patches can be regarded as a soft representation of the object, which can then be fed into the LSTM to generate a proper read key for memory retrieval. However the size of this soft feature map is still too large to directly feed into the LSTM. 
To further reduce the size of each square patch, 
we first adopt an average pooling with $n\times n$ filter size on $F_t$,
\begin{align}
\textbf{f}_t = \text{AvgPooling}_{n\times n}(F_t)
\end{align}
and $\mathbf{f}_{t,i} \in \mathbb{R}^{c}$ is the feature vector 
for the $i$th patch. 

The attended feature vector is then computed as the weighted sum of the feature vectors,
\begin{align}
\mathbf{a}_t = \sum_{i=1}^{L}\alpha_{t,i}\mathbf{f}_{t,i}
\end{align}
where $L$ is the number of square patches, and the attention weights $\alpha_{t,i}$ is calculated by a softmax, 
\begin{align}
\alpha_{t,i} = \frac{\exp(r_{t,i})}{\sum_{k=1}^{L}\exp(r_{t,k})}
\end{align}
where 
\begin{align}
r_{t,i} = W^a \text{tanh}(W^h \mathbf{h}_{t-1}+W^f \mathbf{f}_{t,i}+b)
\end{align}
is an attention network (\ytyy{Figure \ref{fig:3}: Left}), which takes the previous hidden state $\mathbf{h}_{t-1}$ of the LSTM and a square patch $\mathbf{f}^*_{t,i}$ as input. $W^a, W^h, W^f$ and $b$ are weight matrices and biases for the network.

By comparing the target's historical information in the previous hidden state with each square patch, the attention network can generate attentional weights that have higher values on the target and smaller values for surrounding regions.  Figure \ref{fig:3} (right) shows example search images with attention weight maps. Our attention network can always focus on the target which is beneficial when retrieving memory for template matching. 

\subsection{LSTM Memory Controller}

For each time step, the LSTM controller takes the attended feature vector $\mathbf{a}_t$, obtained by the attention module, and the previous hidden state $\mathbf{h}_{t-1}$ as input, and outputs the new hidden state $\mathbf{h}_t$ to calculate the memory control signals, including read key, read strength, bias gates, and decay rate (discussed later).
The internal architecture of the LSTM uses the standard model, while the output layer is modified to generate the control signals.
In addition, we also use layer normalization \cite{Ba2016} and dropout regularization \cite{Srivastava2014} for the LSTM. The initial hidden state $\mathbf{h}_0$ and cell state $\mathbf{c}_0$  
are  
obtained by passing the initial target's feature map through one $n\times n$ average pooling layer and two separate fully-connected layer with tanh activation functions, respectively.

\subsection{Memory Reading}\label{read}

Memory is retrieved by computing a weighted sum of all memory slots with the read weight vector, which is determined by the cosine similarity between the read key and the memory keys. This aims at retrieving the most related template stored in memory. \yty{Since the memory reading processes for positive and negative are similar, we will only show the positive case.}
Suppose $\mathbf{M}_t \in \mathbb{R}^{N\times n \times n \times c}$ represents the memory module,  such that $\mathbf{M}_t(j) \in \mathbb{R}^{n \times n \times c}$ is the template stored in the $j\text{th}$ memory slot and $N$ is the number of memory slots. 
The LSTM controller outputs the read key $\mathbf{k}_t \in \mathbb{R}^{c}$ and read strength $\beta_t \in [1,\infty]$,
\begin{align}
\mathbf{k}_t = & W^k\mathbf{h}_{t}+b^k,  \\
\beta_t = & 1+\log(1+\exp(W^\beta \mathbf{h}_{t}+b^\beta)), 
\end{align}
where 
$W^k, W^\beta, b^k, b^\beta$ are the weight matrices and biases.
The read key $\mathbf{k}_t$ is used for matching the contents in  memory, while the read strength $\beta_t$ indicates the reliability of the generated read key. 
Given the read key and read strength, a \textit{read weight} $\mathbf{w}^r_t\in \mathbb{R}^{N}$ is computed for memory retrieval,
\begin{align}
\mathbf{w}^r_t(j) =\frac{\exp{\{C(\mathbf{k}_t, \mathbf{k}_{\mathbf{M}_t(j)})}\beta_t\}}{\sum_{j'} \exp{\{C(\mathbf{k}_t, \mathbf{k}_{\mathbf{M}_t(j')})}\beta_t\}}, 
\end{align}
where $\mathbf{k}_{\mathbf{M}_t(j)} \in \mathbb{R}^{c}$ is the memory key generated by a $n\times n$ average pooling on $\mathbf{M}_t(j)$. $C(\mathbf{x}, \mathbf{y})$ is the  cosine similarity between vectors, 
$C(\mathbf{x},\mathbf{y})= \frac{\mathbf{x} \cdot \mathbf{y}}{\|\mathbf{x}\|\|\mathbf{y}\|}$.
Finally, the template is retrieved from memory as a weighted sum,
\begin{align}
\mathbf{T}^{\text{retr}}_t=\sum_{j=1}^N\mathbf{w}^r_t(j)\mathbf{M}_t(j).
\end{align}


\subsection{Residual Template Learning}\label{residual}
\begin{figure}[t]
	\begin{center}
		\includegraphics[width=0.95\linewidth]{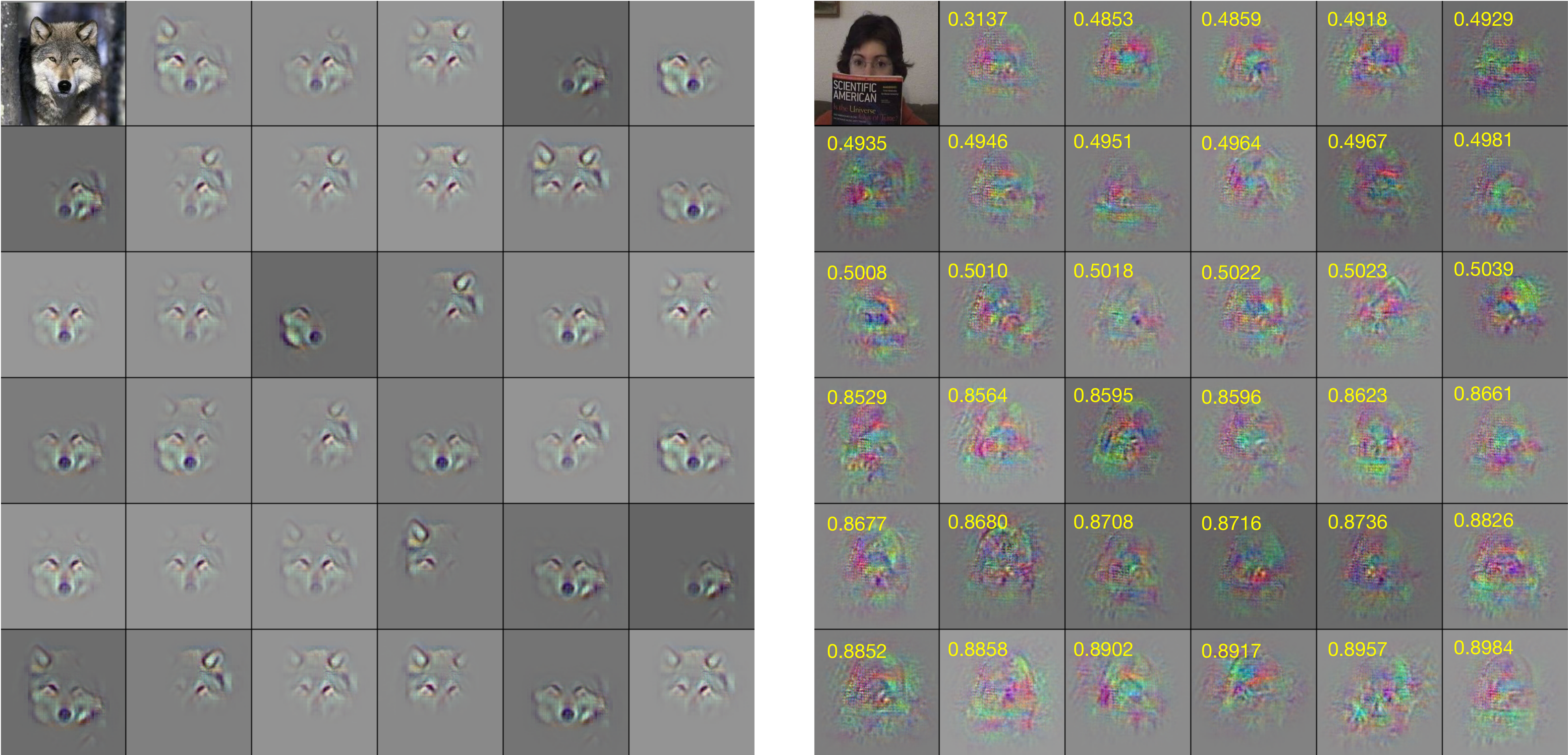}
	\end{center}
	\vspace{-5mm}
	\caption{\textbf{Left}: The feature channels respond to target parts: images are reconstructed from conv5 of the CNN used in our tracker. Each image is generated by accumulating reconstructed pixels from the same channel. The input image is shown in the top-left. \textbf{Right}: Channel visualizations of a retrieved template along with their corresponding residual gate values in the left-top corner.}
	\label{fig:6-1}
	\vspace{-4mm}
\end{figure}

Directly using the retrieved template for similarity matching  is prone to overfit recent frames.
Instead, we learn a residual template by multiplying the retrieved template with a channel-wise gate vector and adding it to the initial template to capture appearance changes. Therefore, our \yty{positive} template is formulated as,
\begin{align}
\mathbf{T}^{\text{pos}}_t = \mathbf{T}_0+ \mathbf{r}_t\odot \mathbf{T}^{\text{retr}}_t,
\end{align}
where $\mathbf{T}_0$ is the initial template and  $\odot$ is channel-wise multiplication.
$\mathbf{r}_t\in \mathbb{R}^c$ is the \textit{residual gate} produced by the LSTM controller, 
\begin{align}
\mathbf{r}_t = \sigma (W^r\mathbf{h}_{t}+b^r),
\end{align}
where $W^r, b^r$ are the weights and bias, and $\sigma$ represents sigmoid function. 
The \textit{residual gate} controls how much each channel of the retrieved template is added to the initial template, which can be regarded as a form of feature selection. 

By projecting different channels of a target feature map to pixel-space using deconvolution, as in \cite{Zeiler2014}, we find that the channels focus on different object parts (Figure \ref{fig:6-1}: Left). 
\ytyy{To show the  behavior of residual learning, we also visualize the retrieved template along with its residual gates in Figure \ref{fig:6-1} (right). The channels that correspond to regions with appearance changes (the bottom part of the face is occluded) have higher residual gate values, demonstrating that the residual learning scheme adapts the initial template to appearance variations. In addition, channels corresponding to previous target appearances are also retrieved from memory (e.g., 6th row, 5th column; the nose and mouth are both visible meaning that they are not occluded), demonstrating that our residual learning does not overfit to recent frames.}
Thus, the channel-wise feature residual learning has the advantage of updating different object parts separately. 
Experiments in Section \ref{abla} show that this yields a big performance improvement. 


\subsection{Distractor Template Canceling and Final Template} 

\begin{figure}[t]
	\begin{center}
		\includegraphics[width=0.9\linewidth]{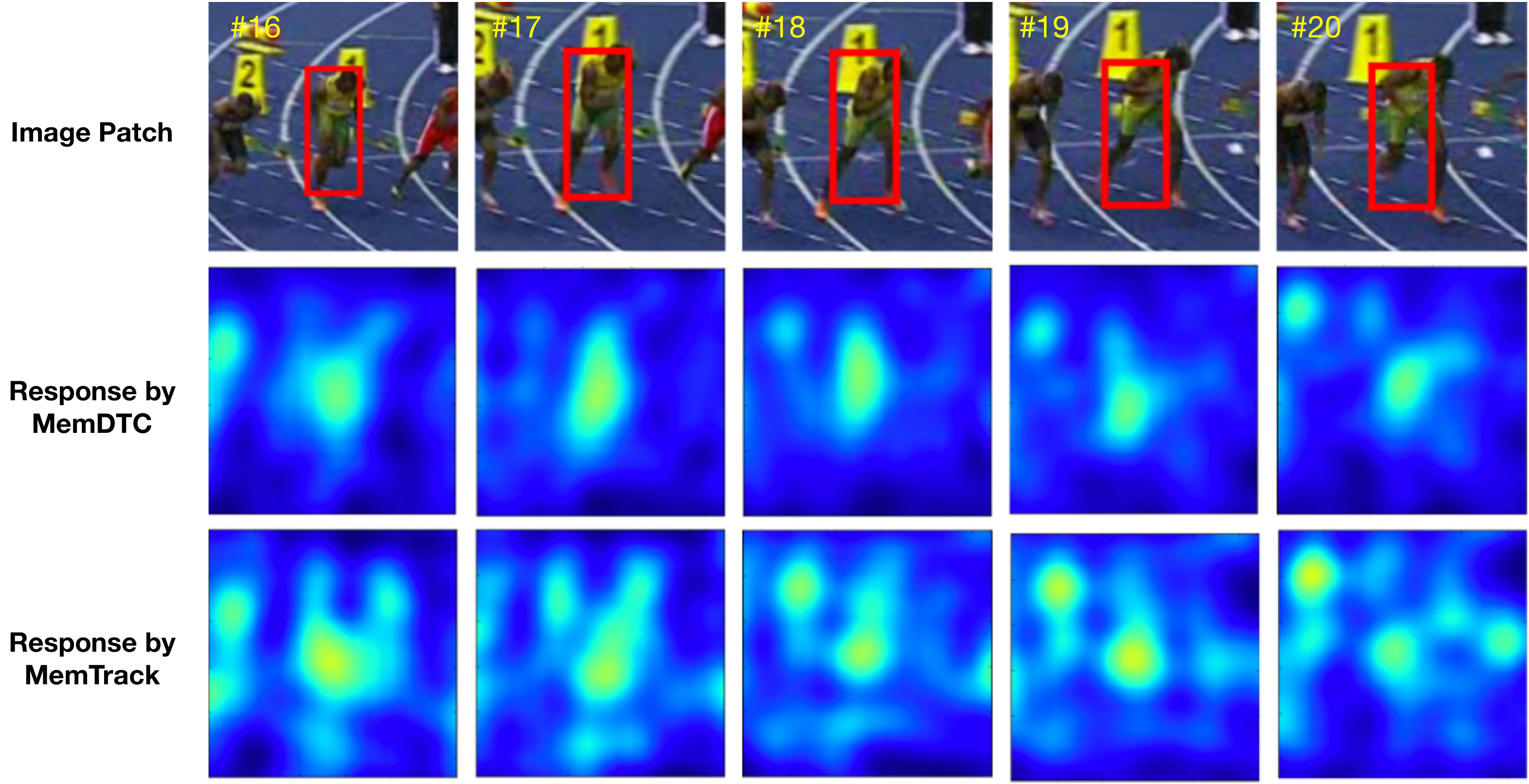}
	\end{center}
	\vspace{-5mm}
	\caption{Example responses comparing tracking with distractor template canceling (MemDTC) and without (MemTrack).}
	\label{fig:6-2}
	\vspace{-4mm}
\end{figure}

\yty{As is shown in Section \ref{residual}, the feature channels respond to different object parts. The channels of the positive template that are similar to those of a distractor template are considered as not discriminative.  Thus we propose to cancel \abc{these} 
feature channels of the positive template via a canceling gate \abc{to obtain the final template,}
\begin{align}
	\mathbf{T}^{\text{final}}_t = \mathbf{T}^{\text{pos}}_t- \mathbf{c}_t\odot \mathbf{T}^{\text{neg}}_t,
\end{align}
where $\mathbf{T}^{\text{neg}}_t$ is the distractor (negative) template which is retrieved from negative memory (as in Section \ref{read}),  and $\mathbf{c}_t$ is the canceling gate produced by comparing the positive and negative templates,
\begin{align}
\mathbf{c}_t = \sigma (W^c\text{tanh}(W^{pos}_{1\times 1\times c}*\mathbf{T}^{\text{pos}}_t+W^{neg}_{1\times 1\times c}*\mathbf{T}^{\text{neg}}_t+b^c))
\end{align}
where 
$W^{pos}, W^{neg}$ are \abc{$1\times 1\times c$ convolution filters}, $\{W^c, b^c\}$ are the weights and bias, and $*$ is the convolution operation. This process weakens the weight of non-discriminative channels when forming the final response map, leading to an emphasis of discerning channels.} \ytyy{To demonstrate the effect of  distractor template canceling, we show the responses generated with distractor template canceling (MemDTC) and without it (MemTrack) 
in Figure \ref{fig:6-2}. The response maps generated by MemDTC are less cluttered than those of MemTrack, and MemDTC effectively suppresses responses from similar looking distractors (i.e., the other runners).}

\tyy{The response map is generated by convolving the search feature map with the final template as the filter, which is equivalent to calculating the correlation score between the template and each translated sub-window on the search feature map in a sliding window fashion. The displacement of the target from the last frame to current frame is calculated by multiplying the position of the maximum score, relative to the center of the response map, with the feature stride. The size of the bounding box is determined by searching at multiple scales of the feature map. This is done using a single feed forward computation by assembling all scaled images as a mini-batch, which is very efficient in modern deep learning libraries. }

\subsection{Positive Memory Writing} \label{memwrite}

\yty{The image patch with the new position of the target is used for positive memory writing.}
The new object template $\mathbf{T}^{\text{new}}_t$ is computed using the feature extraction CNN. There are three cases for memory writing: 1) when the new object template is not reliable (e.g.\ contains a lot of background), there is no need to write new information into memory; 2) when the new object appearance does not change much compared with the previous frame, the memory slot that was previously read should be updated; 
3) when the new target has a large appearance change, a new memory slot should be overwritten.
To handle these three cases, we define the \textit{write weight} as
\begin{align}
\mathbf{w}^w_t =g^s\mathbf{0}+g^r\mathbf{w}^r_t + g^a\mathbf{w}^a_t, 
\end{align}
where $\mathbf{0}$ is the zero vector, $\mathbf{w}^r_t$ is the read weight, and $\mathbf{w}^a_t$  is the allocation weight, which is responsible for allocating a new position for memory writing. 
The \tyy{skip gate} $g^s$, read gate $g^r$ and allocation gate $g^a$, are produced by the LSTM controller with a softmax function, 
\begin{align}
[g^s, g^r, g^a] = \text{softmax}(W^g \mathbf{h}_{t}+b^g),
\end{align}
where $W^g, b^g$ are the weights and biases. Since $g^s+g^r+g^a=1$, these three gates govern the interpolation between the three cases.  If $g^s=1$, then $\mathbf{w}^w_t=\mathbf{0}$ and nothing is written. 
 If $g^r$ or $g^a$ have higher value, then the new template is either used to update the old template (using $\mathbf{w}^r_t$) or written into newly allocated position (using $\mathbf{w}^a_t$). The \textit{allocation weight} is calculated by,
\begin{align}\label{alloc}
\mathbf{w}^a_t(j)=
\begin{cases}
1, &\text{if } j=\displaystyle \mathop{\mathrm{argmin}}_{j} \mathbf{w}^u_{t-1}(j)\\
0, &\text{otherwise}
\end{cases}
\end{align}
where $\mathbf{w}^u_t$ is the \textit{access vector},
\begin{align}
\mathbf{w}^u_t = \lambda \mathbf{w}^u_{t-1} + \mathbf{w}^r_t + \mathbf{w}^w_t,
\end{align}
which indicates the frequency of memory access (both reading and writing), and $\lambda$ is a decay factor. Memory slots that are accessed infrequently will be assigned new templates. \yty{As is shown in Figure \ref{fig:1}, our memory network is able to learn the \abc{appropriate behavior for effectively updating or allocating new templates} to handle appearance variations.}

\begin{figure}[t]
	\begin{center}
		\includegraphics[width=0.8\linewidth]{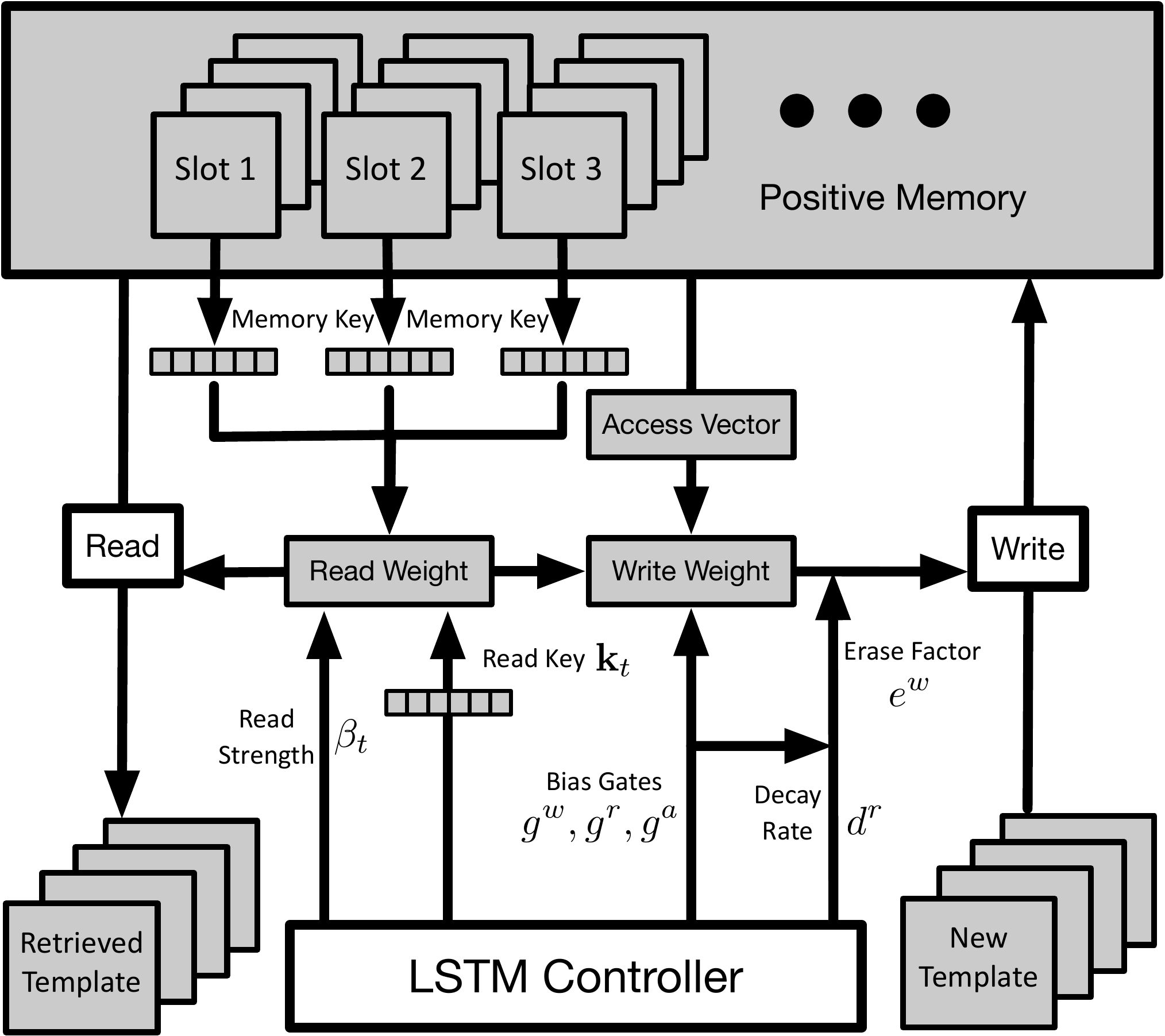}
	\end{center}
	\vspace{-4mm}
	\caption{Diagram of \yty{positive memory access mechanism, including reading and writing process.}
	}
	\vspace{-4mm}
	\label{fig:4}
\end{figure}

The writing process is performed with a \textit{write weight} in conjunction with an \textit{erase factor} for clearing the memory, 
\begin{align}
\mathbf{M}^p_{t+1}(j) = \mathbf{M}^p_{t}(j)(\mathbf{1}-\mathbf{w}^w_t(j)e^w)+\mathbf{w}_t(j)^we^w\mathbf{T}^{\text{new}}_t,
\end{align}
where 
$e^w$ is the \textit{erase factor} computed by
\begin{align}
e^w = d^rg^r+g^a,
\end{align}
and $d^r \in [0,1]$ is the \textit{decay rate} produced by the LSTM controller, 
\begin{align}
d^r = \sigma (W^d\mathbf{h}_{t}+b^d),
\end{align}
where 
$W^d$, $b^d$ are the weights and bias. If $g^r=1$ (and thus $g^a=0$), then $d^r$ serves as the decay rate for updating the template in the memory slot (Case 2). If $g^a=1$ (and $g^r=0$), $d^r$ has no effect on $e^w$, and thus the memory slot will be erased before writing the new template (Case 3). Figure \ref{fig:4} shows the detailed diagram of the positive memory reading and writing process.

\subsection{Negative Memory Writing}
\label{text:negmemwrite}
\yty{For negative memory, the distractor templates for memory writing are extracted from the search feature map based on their response score. Those with high score peak and are far away from the target are considered as \ytyy{distractor} templates. \tyy{Following the notation of Section \ref{attention}, $F_{t,i}$ is the $i$-th $\mathit{n\times n\times c}$ square patch on search feature map $F_t$.}
%
\abc{The set of distractor templates is defined as the top-$K$ templates (based on response score),}
\begin{align}
\begin{split}
{\mathcal S}_{\text{dis}}=\{F_{t,i} \mid D(i,*) >\tau, R(F_{t,i}) > \gamma R(F_{t,*}),  \\
F_{t,i} \in \argmax_{F_t}^K R(F_t) \}, 
\end{split}
\end{align}
\abc{where $R(F_{t,i})$ is the response score for $F_{t,i}$, $D(i,*)$ is the spatial distance between the centers of the two templates, and $F_{t,*}$ is the template with maximum response score.} 
\abc{The operator $\argmax^K$ returns the set with the top-$K$ values.}  $\tau$ is a distance threshold, and $\gamma$ is a score ratio threshold. 
%
Note that if there are no distractor templates satisfying the above criterion, a zero-filled template will be written into the negative memory, which will have no effect on forming the final template.}

\yty{As the distractor template is usually temporary and changes frequently, we simplify the memory writing process by only using an allocation weight as the write weight. Thus the negative memory writing \abc{is similar to a memory queue}
\begin{align}
\mathbf{M}^n_{t+1}(j) = \sum_k^K \mathbf{M}^n_{t}(j)(\mathbf{1}-\mathbf{w}^{na}_{t,k}(j))+\mathbf{w}^{na}_{t,k}(j)\mathbf{T}^{\text{dis}}_{t,k},
\end{align}
where $\mathbf{T}^{\text{dis}}_{t,k} \in \mathcal S_{dis}$ stands for the \textit{k}-th distractor template selected based on response score. $\mathbf{w}^{na}_{t,k}$ is the allocation weight for negative memory, \tyy{which is calculated by
\begin{align}
\mathbf{w}^{na}_{t,k}(j)=
\begin{cases}
1, &\text{if } j=  p(k), p(k)\in \mathcal S_{alloc}  \\
0, &\text{otherwise}
\end{cases}
\end{align}
where 
$\mathcal S_{alloc} = \displaystyle \mathop{\mathrm{argmin}}_{j}^K \mathbf{w}^u_{t-1}(j)$ 
represents the top-$K$ newly allocated memory positions.}} 


\subsection{Auxiliary Classification Loss}
\yty{As is stated in \cite{Ma2015}, robust visual tracking needs both fine-grained details to accurately localize the target and semantic information to be robust to appearance variations caused by deformation or occlusion. Features learned with similarity matching are mainly focused on precise localization of the target. Thus, we propose to add an auxiliary classification branch after the last layer of the CNN feature extractor to guide the networks to learn complementary semantic information. 
The classification branch contains a  fully-connected layer with 1024 neurons and ReLU activations, followed by a fully connected layer with 30 neurons (there are 30 categories in the training data) with a softmax function. The final loss for optimization is formed by two parts, matching loss and classification loss, 
\tyy{\begin{align}
	L(R, R^*, p, p^*) = L_{\text{mch}}(R, R^*) + \kappa L_{cls}(p, p^*),
\end{align}
where $R, R^*$ are the predicted response and groundtruth response. $p, p^*$ are the predicted probability and the groundtruth class of the object. $L_{\text{mch}}$ is an element-wise sigmoid cross entropy loss as in \cite{Bertinetto2016},
\begin{align}
L_{\text{mch}}(R, R^*)  = \frac{1}{|\mathcal{D}|}\sum_{u\in \mathcal{D}} \ell(R_u, R_u^*),
\end{align}
where $\mathcal{D}$ are the positions   
in the score map, and $\ell(\cdot)$ is the sigmoid cross entropy loss. 
$L_{cls}$ is the softmax cross entropy loss. $\kappa$ is a balancing factor between the two losses. Note that the classification branch will be removed during testing.}
}


\section{Implementation Details}
We adopt an Alex-like CNN as in SiamFC \cite{Bertinetto2016} for feature extraction, where the input image sizes of the object  and search images are 127$\times$127$\times$3 and 255$\times$255$\times$3, respectively. \tyy{We use the same strategy for cropping the search and object images as \cite{Bertinetto2016}, where some context margins around the target are added when cropping the object image. Specifically, given the newly predicted bounding box $\{x_t, y_t, w_t, h_t\}$ (center x, center y, width, height) in frame $t$,  the cropping ROI for the object image patch is calculated by
\begin{align}
	x_t^o &= x_t, \quad y_t^o = y_t,  \quad
	w_t^o = h_t^o  = \sqrt{(c+w_t)(c+h_t)}, 
\end{align}
where $c = \delta*(w_t+h_t)$ is the context length and $\delta = 0.5$ is the context factor. For frame $t+1$, the ROI cropping for the search image patch is computed by
 \begin{align}
 \begin{split}
 x_{t+1}^s &= x_t, \quad y_{t+1}^s = y_t, \\
 w_{t+1}^s &= h_{t+1}^s  = \frac{255-127}{127}*w_t^o+w_t^o.
\end{split}
\end{align}
Note that the cropped object image patch and search image patch are then resized to 127$\times$127$\times$3 and 255$\times$255$\times$3, respectively, to match the network input. }

The whole network is trained offline on the VID dataset (object detection from video) of ILSVRC \cite{ILSVRC15} from scratch, and takes about one day. 
Adam \cite{kingma2014adam} optimization is used with a mini-batches of 8 video clips of length 16. The initial learning rate is 1e-4 and is multiplied by 0.8 every 10k iterations. The video clip is constructed by 
uniformly sampling frames (while keeping the temporal order) from each video. This aims to diversify the appearance variations in one episode for training, which can simulate fast motion, fast background change, jittering object, low frame rate.
We use data augmentation, including small image stretch and translation for the target image and search image. 
The dimension of memory states in the LSTM controller is 512 and the retain probability used in dropout for LSTM is 0.8. \yty{The number of positive memory slots and negative memory slots are $N_{pos}=8, N_{neg}=16$. The distance threshold and score ratio threshold are $\tau = 4, \gamma=0.7$ and the number of selected \ytyy{distractor} templates is $K=2$.  The balancing factor for auxiliary loss is $\kappa = 0.05$.} The decay factor used for calculating the access vector is $\lambda=0.99$.
%

At test time, the tracker runs completely feed-forward and no online fine-tuning is needed. We locate the target based on the upsampled response map as in SiamFC \cite{Bertinetto2016}, and handle the scale changes by searching for the target over three scales $1.05^{[-1,0,1]}$. To smoothen scale estimation and penalize large displacements, we update the object scale with the new one by an exponential smoothing factor of 0.5 and dampen the response map with a cosine window by an exponential smoothing factor of 0.19.

Our algorithm is implemented in Python with the TensorFlow toolbox \cite{abadi2016tensorflow}, and tested on a computer with four Intel(R) Core(TM) i7-7700 CPU @ 3.60GHz and a single NVIDIA GTX 1080 Ti with 11GB RAM. \yty{It runs about 50 fps for MemTrack and MemTrack* and about 40 fps for MemDTC and MemDTC*.}

\section{Experiments}

We evaluate our preliminary tracker \cite{Yang2018}
which only has positive memory networks (MemTrack), as well as three improved versions which are MemTrack with auxiliary classification loss (MemTrack*), MemTrack with \ytyy{distractor} template canceling (MemDTC), and MemDTC with auxiliary classification loss (MemDTC*). We conduct experiments on five challenging datasets: OTB-2013 \cite{Wu2013}, OTB-2015 \cite{Wu2015}, VOT-2015 \cite{Kristan2015}, VOT-2016 \cite{Kristan2016} and VOT-2017 \cite{Kristan2017}.  We follow the standard protocols, and evaluate using precision and success plots, as well as area-under-the-curve (AUC) on OTB datasets. We also present the distance precision rate, overlap success rate and center location error on OTB for completeness. For VOT datasets, we use the toolbox\footnote{\href{https://github.com/votchallenge/vot-toolkit}{https://github.com/votchallenge/vot-toolkit}}  provided by VOT committee to generate the results.

\subsection{OTB datasets}

On OTB-2013 and OTB-2015, we compare our proposed trackers with 12 recent {\em real-time} methods ($\geq$ 15 fps): \ytyy{SiamRPN \cite{Li2018}, DSiamM \cite{Guo2017}, PTAV \cite{Fan2017}}, CFNet \cite{Valmadre2017}, LMCF \cite{Wang2017}, ACFN \cite{Choi2017}, RFL \cite{Yang2017}, SiamFC \cite{Bertinetto2016}, SiamFC* \cite{Valmadre2017}, Staple \cite{Bertinetto2016-1}, DSST \cite{Danelljan2014}, and KCF \cite{Henriques2015} . To further show our tracking accuracy, on OTB-2015, we also compared with another 8 recent state-of-the art trackers that are {\em not} real-time speed: CREST \cite{Song2017},  CSR-DCF \cite{Lukezic2017}, MCPF \cite{Zhang2017}, SRDCFdecon \cite{Danelljan2016}, SINT \cite{Tao2016}, SRDCF \cite{Danelljan2015}, HDT \cite{Qi2016}, HCF \cite{Ma2015}.

The OTB-2013 \cite{Wu2013} dataset contains 51 sequences with 11 video attributes and two evaluation metrics, which are center location error and overlap ratio. The OTB-2015 \cite{Wu2015} dataset is the extension of OTB-2013 to 100 sequences, and is thus more challenging. We conduct the ablation study mainly on OTB-2015 since it contains OTB-2013.

\begin{figure}[t]
	\begin{center}
		\includegraphics[width=\linewidth]{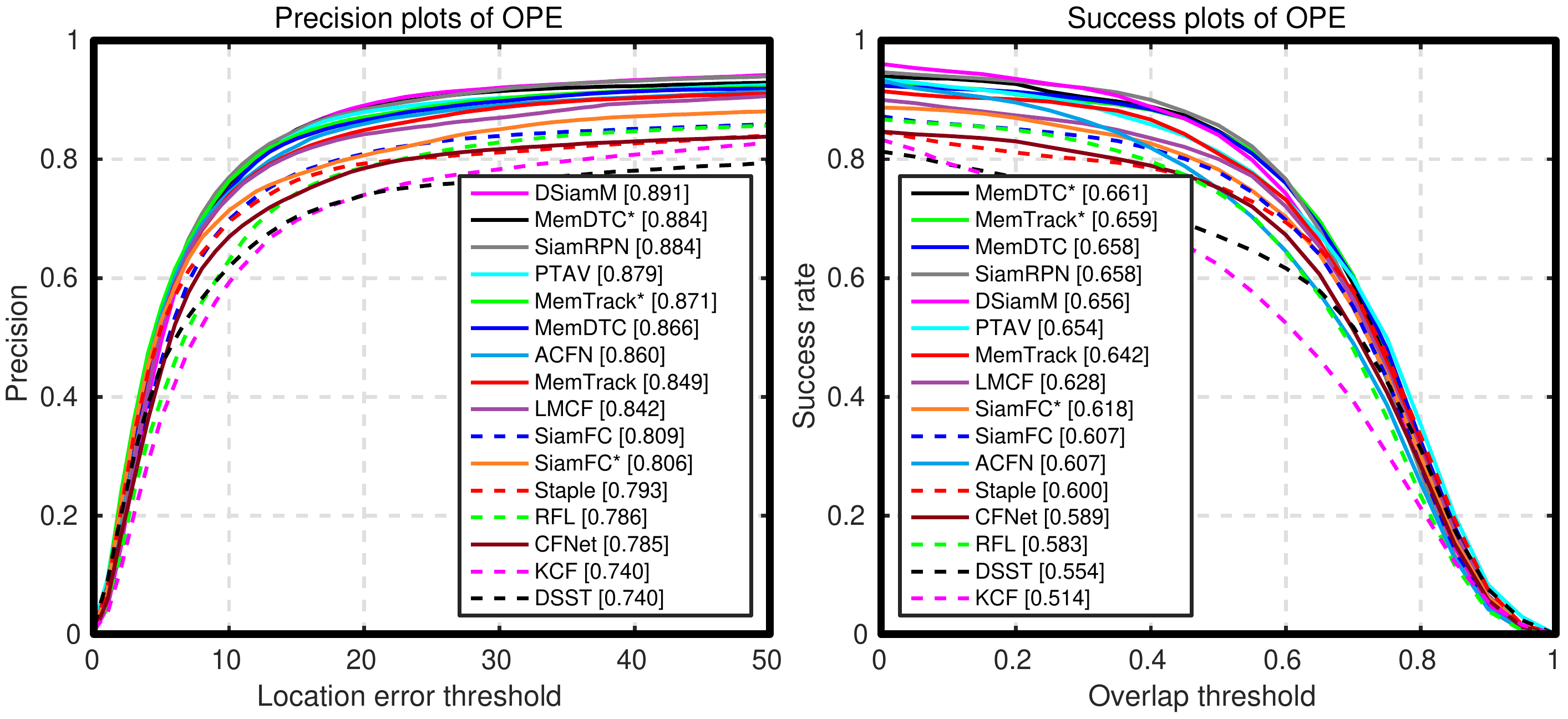}
	\end{center}
	\vspace{-5mm}
	\caption{Precision and success plots on OTB-2013 for  real-time trackers.}
	\vspace{-4mm}
	\label{fig:8}
\end{figure}

\begin{figure}[t]
	\begin{center}
		\includegraphics[width=\linewidth]{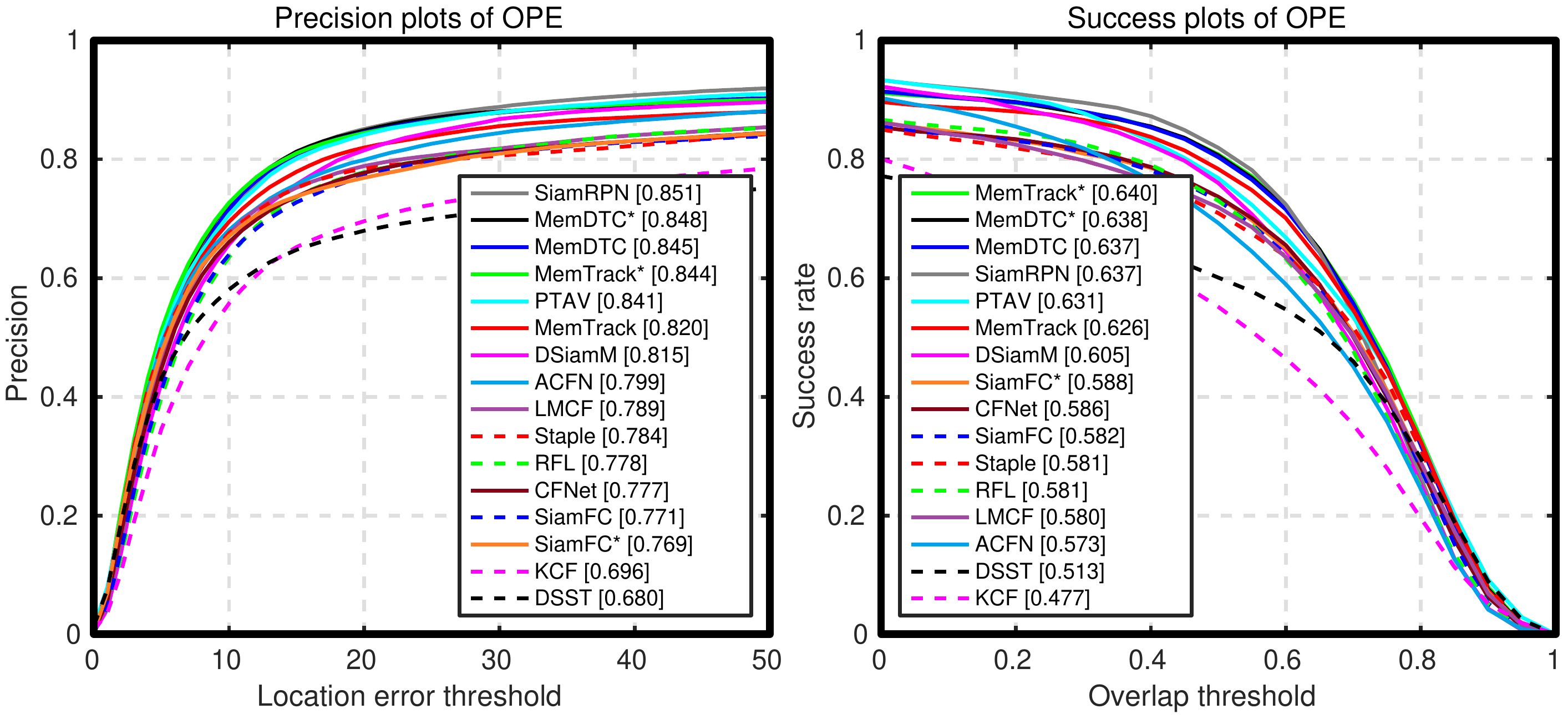}
	\end{center}
	\vspace{-5mm}
	\caption{Precision and success plot on OTB-2015 for real-time trackers.}
	\label{fig:9}
	\vspace{-5mm}
\end{figure}

\begin{table*}
	\begin{center}
		\bgroup
		\def\arraystretch{1.25}
		\resizebox{0.95\textwidth}{!}{
		\begin{tabular}{ccccccccccccccccccc}
			\hline 
				 & & MemDTC* &MemDTC & MemTrack* & MemTrack & SiamRPN& DSiamM& PTAV& LMCF & ACFN & SiamFC &SiamFC* & RFL & Staple & CFNet & KCF & DSST \\
			\hline
			\multirow{2}{*}{DP @10 (\%) ($\uparrow$)} & I & 75.9 &74.8 & \underline{76.4} & 73.9 & \textbf{76.9}&75.8&74.7&73.6 & 74.4 & 69.8 & 71.4 & 63.0 & 69.5 & 66.9 & 59.2 & 61.9\\
			& II &\underline{72.5} &71.6 & \textbf{72.7} & 69.5 &71.3&65.7&70.7& 67.4 & 68.1 & 63.9 & 67.1 & 63.5 & 67.6 & 66.0 & 55.7 & 58.1\\
			\multirow{2}{*}{DP @20 (\%) ($\uparrow$)} & I &\underline{88.4} & 86.6 & 87.1 & 84.9 & \underline{88.4}&\textbf{89.1}&87.9&84.2 & 86.0 & 80.9 & 80.6 & 78.6 & 79.3 & 78.5 & 74.0 & 74.0\\
			& II &\underline{84.8} & 84.5 & 84.4 & 82.0 &\textbf{85.1}&81.5&84.1& 78.9 & 79.9 & 77.1 & 76.9 & 77.8 & 78.4 & 77.7 & 69.6 & 68.0\\
			\multirow{2}{*}{DP @30 (\%) ($\uparrow$)} & I &91.5 & 89.7 & 90.2 & 88.7 & \textbf{92.0}&\underline{91.7}&90.3&87.0 & 89.1 & 83.9 & 85.0 & 82.8 & 81.1 & 81.7 &78.3 & 76.4\\
			& II &\underline{88.1} & 87.8 & 87.9 & 85.6 &\textbf{88.8}&86.8&88.0& 81.8 & 84.4 & 81.1 & 80.9 & 81.7 & 80.6 & 81.3 & 74.0 & 71.3\\
			\hline
			\multirow{2}{*}{OS @0.3 (\%) ($\uparrow$)} & I&90.5 &90.2 & 89.7  & 88.9 & \textbf{92.0}&\underline{92.0}&89.3&86.0 & 86.5 & 83.9 & 84.9 & 82.9 &80.0 &81.0 &73.0 &74.5\\
			& II& 87.7 & \underline{88.0} & 87.9  & 86.6 &\textbf{89.5}&86.3&87.8& 79.8 & 81.9 & 81.2 & 80.9 & 82.5 &79.8 &81.5 &68.1 &69.1\\
			\multirow{2}{*}{OS @0.5 (\%) ($\uparrow$)} & I& \underline{84.7}& 84.6 & 84.5  & 80.9 & \textbf{85.7}&84.1&81.3&80.0 & 75.0 & 77.9 & 78.3 & 74.3 &75.4 &75.2 &62.3 &67.0\\
			& II&80.6 & 80.3 & \textbf{80.8}  & 78.3 &\textbf{81.9}&76.0&76.8& 71.9 & 69.2 & 73.0 & 73.6 & 73.0 &70.9 &73.7 &55.1 &60.1\\
			\multirow{2}{*}{OS @0.7 (\%) ($\uparrow$)} & I&59.0 & \underline{60.0} & \textbf{60.4}  & 57.3 & 56.8&56.1&59.9&56.5 & 49.0 & 55.1 & 55.0 & 48.0 &57.9 &51.3 &39.3 &51.8\\
			& II&55.7 & \underline{55.7} & \textbf{56.2}  & 54.6 &53.9&48.5&53.4& 50.2& 45.1 & 50.3 & 50.8 & 47.7 &51.7 &50.1 &35.3 &46.0\\
			\hline
			\multirow{2}{*}{CLE (pixel) ($\downarrow$)} & I&16.9& 21.5 & 19.1 &  27.6 &\underline{14.2}&\textbf{13.8}&19.3& 23.8 & 18.7 & 29.7 & 35.2 & 35.7 &30.6 &40.3 &35.5&41.4\\
			& II&\underline{20.3} & 21.8 & 22.1 &  27.8 &\textbf{19.2}&22.8&19.8& 39.0 & 25.2 & 33.1 & 35.9 & 35.8 &31.4 & 34.8 &44.7 &50.3\\
			\hline
		\end{tabular}}
		\egroup
	\end{center}
	\caption{Comparison results on OTB-2013 (I) and OTB-2015 (II).  DP @$n$ is the distance precision rate at the threshold of $n$ pixels and OS @$s$ is the overlap success rate at the threshold of $s$ overlap ratio. CLE is center location error.  The best result is bolded, and second best is underlined. The up arrows indicate higher values are better for that metric, while down arrows mean lower values are better.}
	\label{tb:1}
	\vspace{-9mm}
\end{table*}

\subsubsection{Comparison to real-time trackers} 
Figure \ref{fig:8} shows the one-pass comparison results with recent real-time trackers on OTB-2013. Our newly proposed trackers MemDTC, MemTrack* and MemDTC* rank the three best AUC scores on the success plot, which all outperform our earlier work MemTrack \cite{Yang2018}. For the precision plot with center location error, these three trackers also surpass MemTrack by a large margin.
Compared with SiamFC \cite{Bertinetto2016}, which is the baseline for matching-based methods without online updating, the proposed MemDTC*, MemDTC and MemTrack* achieve an improvement of 9.3\%, 7.0\% and 7.7\% on the precision plot, and 8.9\%, 8.4\% and 8.6\% on the success plot. 
Our methods also outperforms SiamFC*, the improved version of SiamFC \cite{Valmadre2017} that uses 
linear interpolation of the old and new filters with a small learning rate for online updating. 
This indicates that our dynamic memory networks can handle object appearance changes better than simply interpolating new templates with old ones.

Figure \ref{fig:9} presents the precision plot and success plot for recent real-time trackers on OTB-2015. Our newly proposed trackers outperform all other methods \ytyy{on success plot with AUC score.} Specifically, our methods perform much better than RFL \cite{Yang2017}, which uses the memory states of the LSTM to maintain the object appearance variations. This demonstrates the effectiveness of using an external addressable memory to manage object appearance changes, compared with using LSTM memory which is limited by the size of the hidden states.
Furthermore, the proposed MemDTC*, MemDTC and MemTrack* improve over the baseline 
SiamFC \cite{Bertinetto2016} by 10.0\%, 9.6\% and 9.5\% on the precision plot, and 9.6\%, 9.5\% and 10.0\% on the success plot.
 Our trackers \ytyy{MemTrack* and MemDTC*} also outperform the most recently proposed three trackers,  \ytyy{SiamRPN \cite{Li2018}, DSiamM \cite{Guo2017}, PTAV \cite{Fan2017}}, on AUC score. 

Table \ref{tb:1} shows the quantitative results of the distance precision (DP) rate at different pixel thresholds (10, 20, 30), overlap success (OS) rate at different overlap ratios (0.3, 0.5, 0.7), and center location errors (CLE) on both OTB-2013 (I) and OTB-2015 (II).  \ytyy{Our improved trackers MemDTC*, MemDTC and MemTrack* consistently outperform our earlier work MemTrack \cite{Yang2018} on all measures. In addition, they also performs well when the success condition is more strict (DP @10 and OS @0.7), indicating that their estimated bounding boxes are more accurate.}

\begin{figure*}[t]
	\begin{center}
		\includegraphics[width=\linewidth]{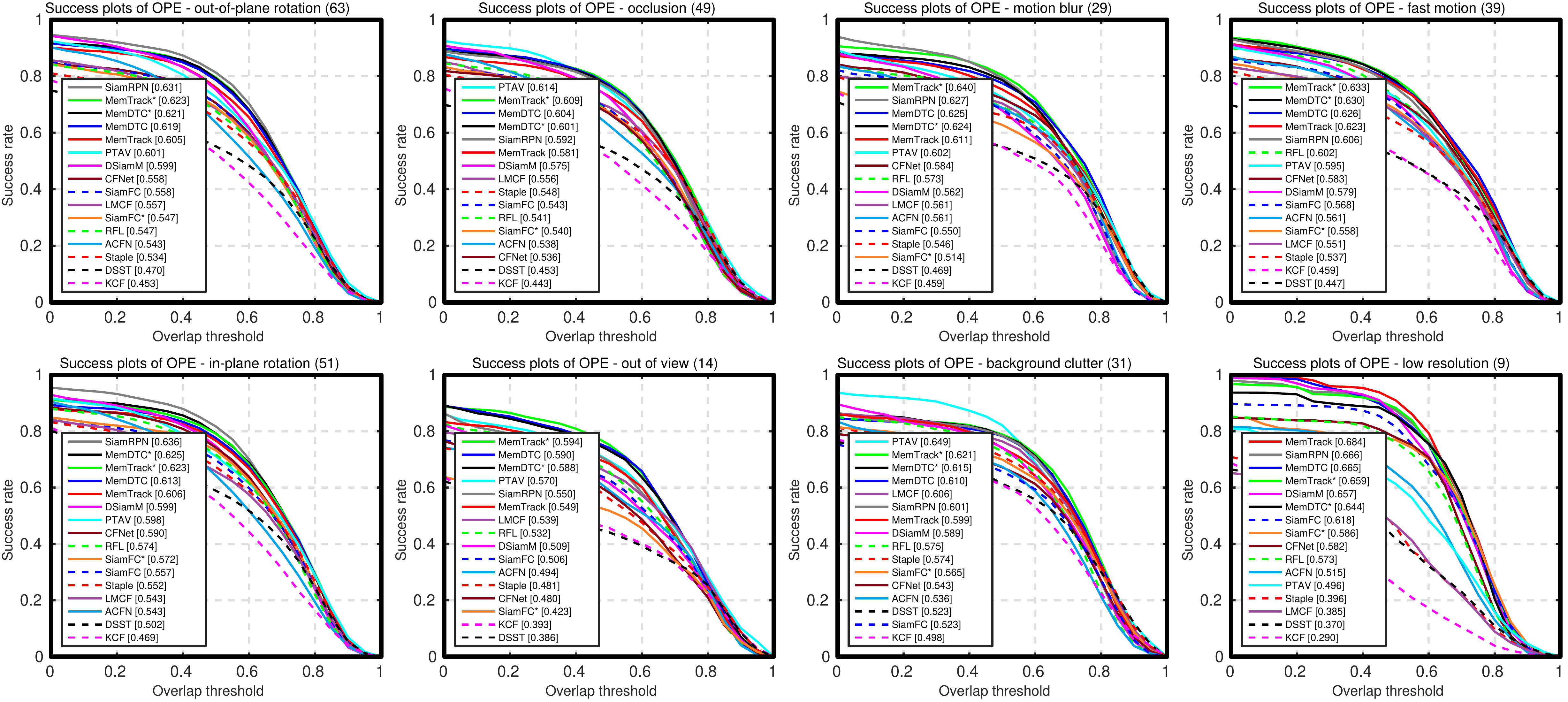}
	\end{center}
	\vspace{-1mm}
	\caption{The success plots on OTB-2015 for eight challenging attributes: out-of-plane rotation, occlusion, motion blur, fast motion, in-plane rotation, out of view, background clutter and low resolution.}
	\label{fig:11}
	\vspace{-1mm}
\end{figure*}

\begin{figure*}[t]
	\begin{center}
		\includegraphics[width=0.95\linewidth]{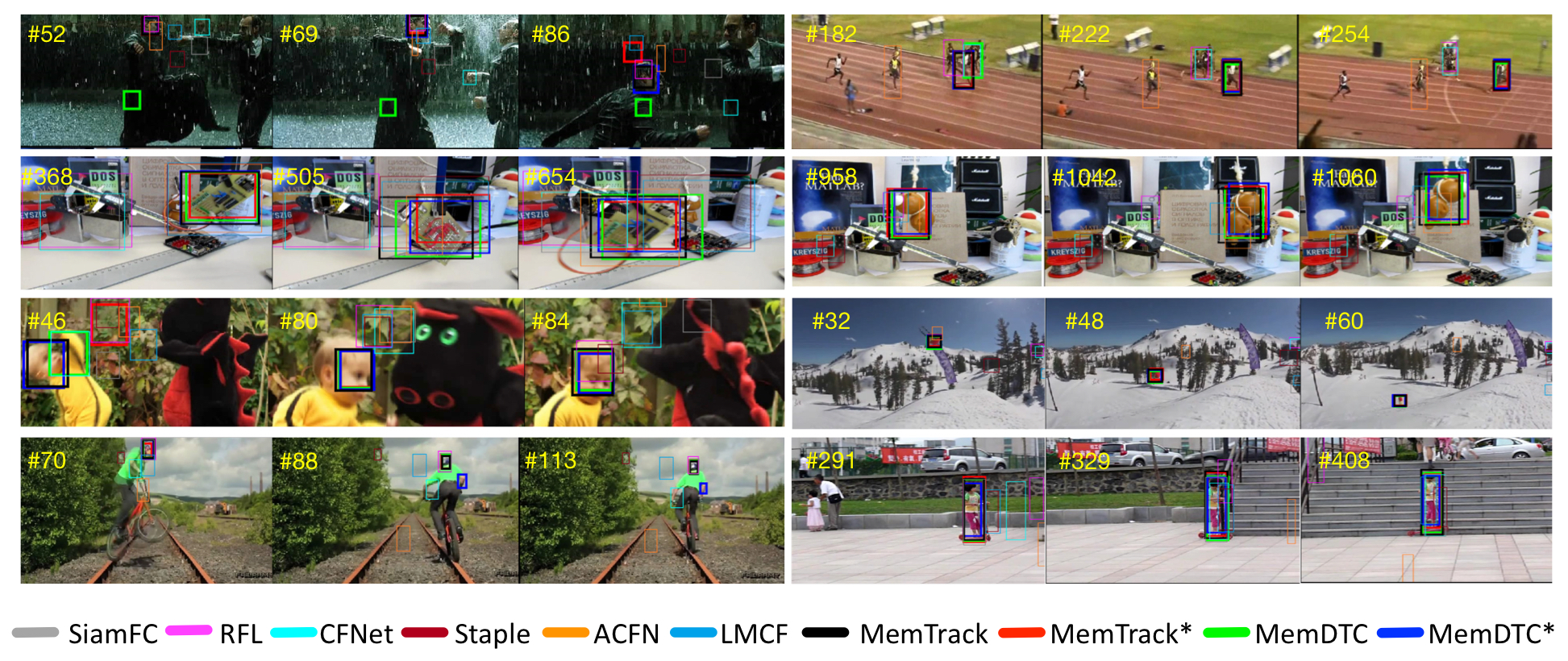}
	\end{center}
	\vspace{-1mm}
	\caption{Qualitative results of our MemTrack, along with  SiamFC \cite{Bertinetto2016}, RFL \cite{Yang2017}, CFNet \cite{Valmadre2017},  Staple \cite{Bertinetto2016-1}, LMCF \cite{Wang2017}, ACFN \cite{Choi2017} on eight challenge sequences. From left to right, top to bottom: \textit{board, bolt2, dragonbaby, lemming, matrix, skiing, biker, girl2}.}
	\label{fig:12}
	\vspace{-1mm}
\end{figure*}

Figure \ref{fig:11} further shows the AUC scores of real-time trackers on OTB-2015 under different video attributes including out-of-plane rotation, occlusion, motion blur, fast motion, in-plane rotation, out of view, background clutter and low resolution. \ytyy{Our MemTrack* outperforms other trackers on  motion blur, fast motion and out of view, demonstrating its ability of adapting to appearance variations using multiple memory slots.  In addition, our MemTrack also shows superior accuracy on low-resolution attributes.} 
Figure~\ref{fig:12} shows qualitative results of our trackers compared with 6 real-time trackers.  

\subsubsection{Comparison to non-real-time trackers} 
\begin{figure}[t]
	\begin{center}
		\includegraphics[width=\linewidth]{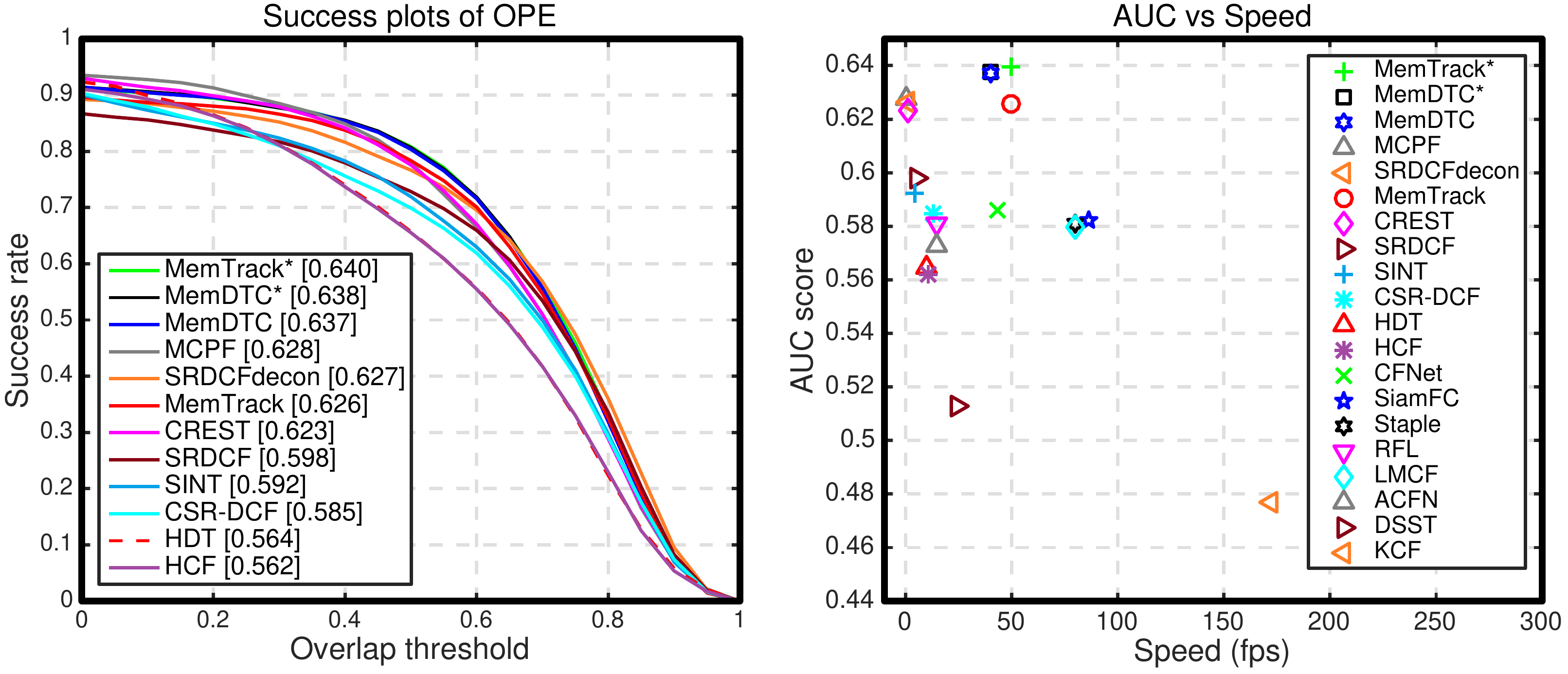}
	\end{center}
	\vspace{-5mm}
	\caption{(left) Success plot on OTB-2015 comparing our real-time methods with recent {\em non-real-time} trackers. (right) AUC score vs.~speed with recent trackers.}
	\label{fig:10}
	\vspace{-5mm}
\end{figure}

 Figure \ref{fig:10} presents the comparison results of 8 recent state-of-the-art {\em non-real time} trackers for AUC score (left), and the AUC score vs.~speed (right) of all trackers. Our newly proposed trackers MemDTC*, MemDTC and MemTrack*, which run in real-time (40 fps for MemDTC* and MemDTC, 50 fps for MemTrack*), outperform CREST \cite{Song2017}, MCPF \cite{Zhang2017} and SRDCFdecon \cite{Danelljan2016}, which all run at $\sim$1 fps. 
Moreover, our earlier work MemTrack also surpasses SINT, which is another matching-based method with optical flow as motion information, in terms of both accuracy and speed.

\subsection{VOT datasets}

The VOT-2015 dataset \cite{Kristan2015} contains 60 video sequences with per-frame annotated visual attributes. Objects are marked with rotated bounding boxes to better fit their shapes. The VOT-2016 dataset \cite{Kristan2016} uses the same sequences as in VOT-2015 but re-annotates the ground truth bounding boxes in an automatic way with per-frame segmentation masks. The VOT-2017 dataset \cite{Kristan2017} replaces the least challenging sequences in VOT-2016 with 10 new videos and manually fixes the bounding boxes that  were incorrectly placed by automatic methods introduced in VOT-2016.

Tracker performance is evaluated using three metrics: expected average  overlap (EAO), accuracy, and robustness. The expected average overlap is computed by averaging the average overlap on a large set of sequence clips with different predefined lengths for all videos. The accuracy measures how well the bounding box estimated by the tracker fits the ground truth box and the robustness measures the frequency of tracking failure during tracking. 

\subsubsection{VOT-2015 Results} 
\begin{figure}[t]
	\begin{center}
		\includegraphics[width=0.47\linewidth]{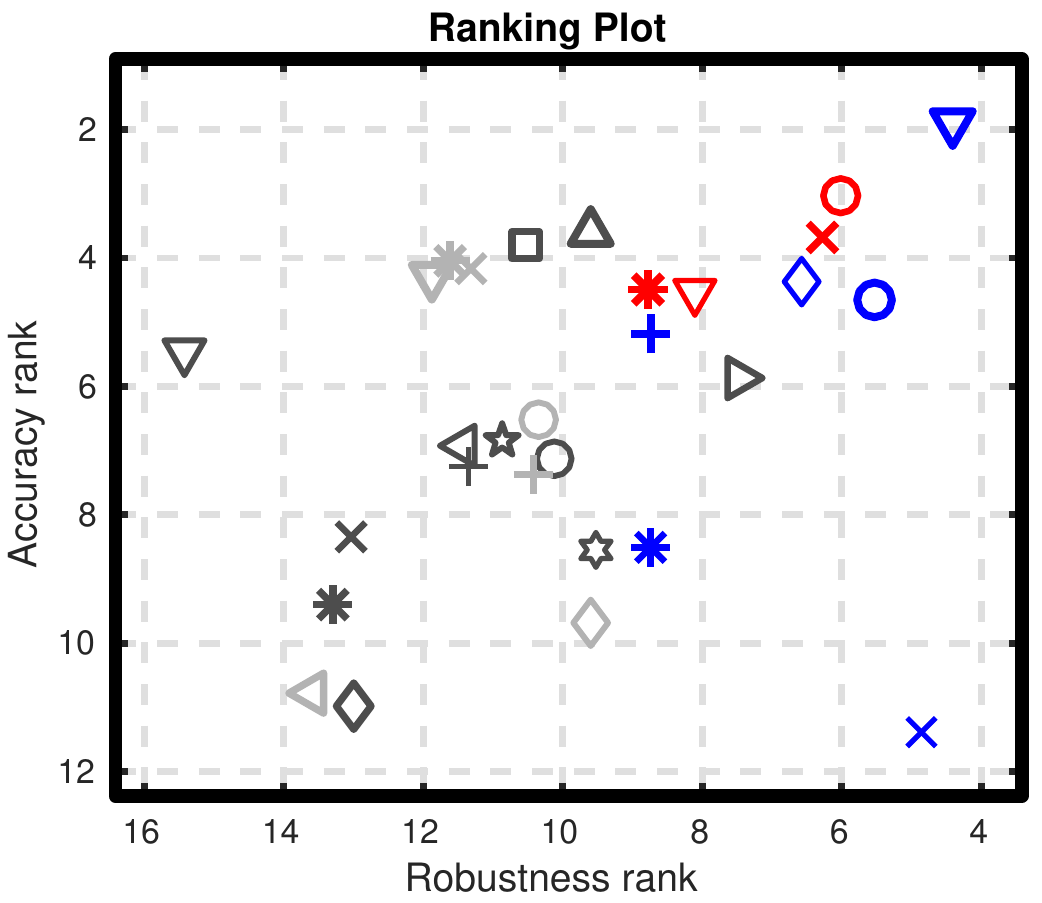}
		\includegraphics[width=0.47\linewidth]{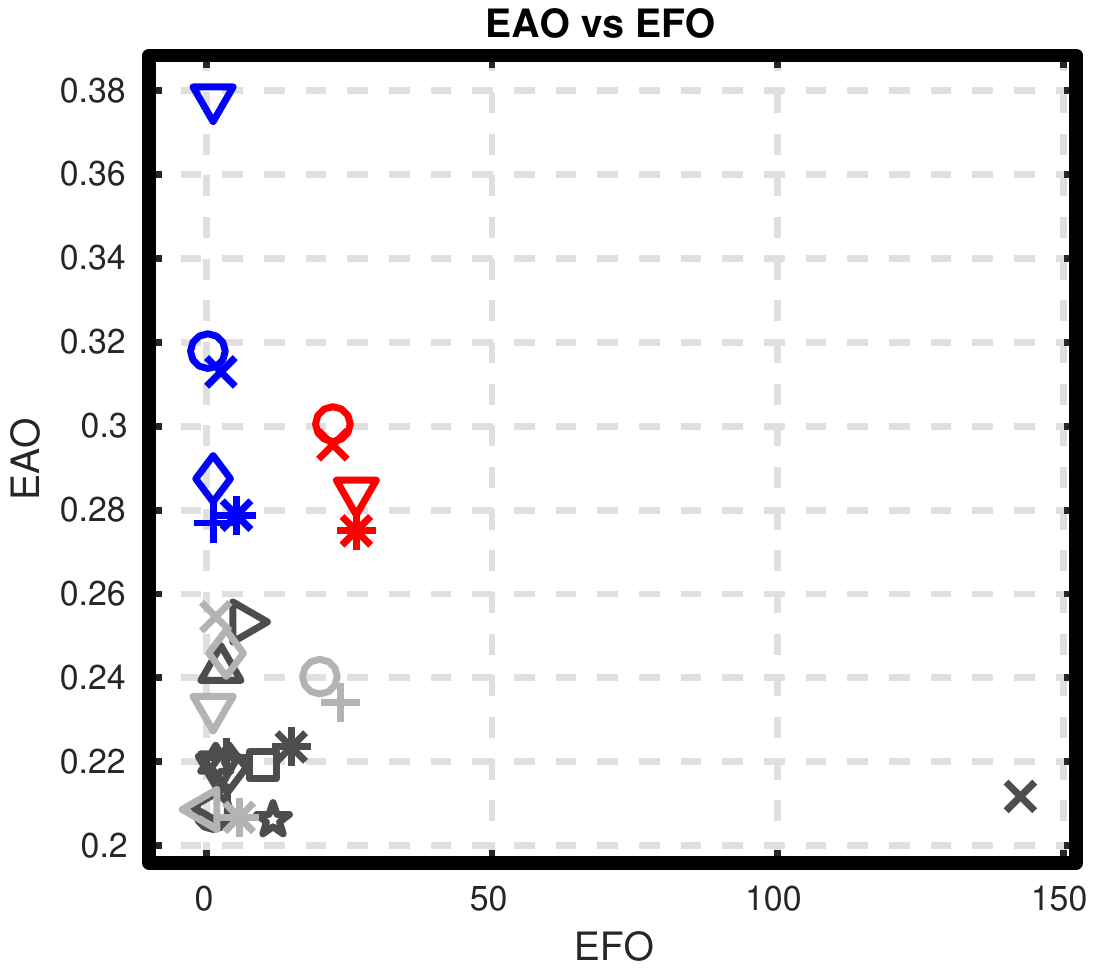}
		\includegraphics[width=0.9\linewidth]{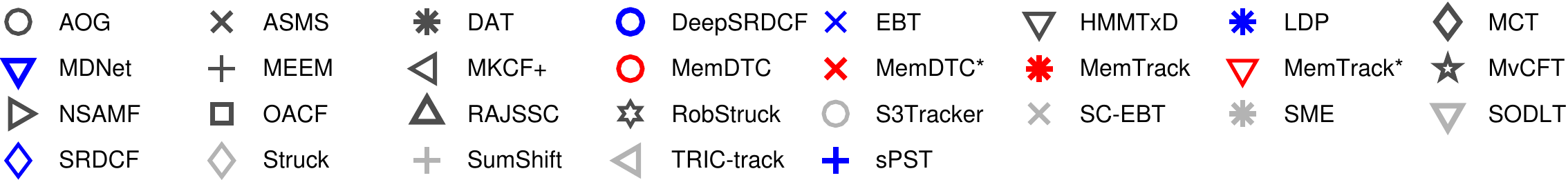}
	\end{center}
	\vspace{-3mm}
	\caption{The AR rank plots and \abcn{EAO vs. EFO} 
	for VOT-2015. Our methods are colored with red, while the top-10 methods are marked with blue. Others are colored with gray.
	}
	\vspace{-2mm}
	\label{fig:13}
\end{figure}

There are 41 trackers submitted to VOT-2015 and 21 baseline trackers contributed by the VOT-2015 committee. Table \ref{tb:2} presents the detailed comparison results with the best 25 performing methods \abcn{(according to EAO, expected average overlap)}.
 Our newly proposed tracker MemDTC* achieves the third and fourth place in terms of accuracy and EAO. 
 Note that MDNet \cite{Nam2016} achieves the best score for all metrics, which is however much slower than our MemTrack*. Our methods also runs much faster (higher EFO\footnote{\tyyy{EFO (equivalent filter operations) is a measurement of speed generated by the VOT toolkit automatically, and is similar to fps but is a relative value. For example, SiamFC gets 32 EFO (VOT) vs. 86 fps (original) in Table \ref{tb:4}, while ours is 24 EFO (VOT) vs.~50 fps (original).}}) than DeepSRDCF \cite{Danelljan2016-2} and EBT \cite{Zhu2016}, which are slightly better than our MemTrack*. Moreover, SRDCF \cite{Danelljan2015}, LDP \cite{Kristan2015} and sPST \cite{Hua2015}, which outperform our MemTrack, do not have real-time speed. Figure \ref{fig:13} shows the accuracy-robustness ranking plot (left) and EAO vs.~EFO 
  plot (right) on the VOT-2015 dataset. Our methods perform favorably against state-of-the-art trackers in terms of both accuracy and robustness (see upper right corner), \abcn{while maintaining real-time speed ($\sim$20 EFO).} \tyyy{Finally, MemTrack and MemTrack* have slightly worse EAO than the baseline SiamFC on VOT-2015, which could be an artifact of the noisy ground-truth bounding boxes in VOT-2015. 
  On VOT-2016, which contains the same videos as VOT-2015 and has improved ground-truth annotations, MemTrack and MemTrack* outperform SiamFC by a large margin  (see Section 5.2.2).}
 

\begin{table}
\small
\begin{center}
\bgroup
\def\arraystretch{1.15}
\begin{tabular}{ccccc }
	\hline
	\multirow{2}{*}{\textbf{ }}
	& EAO ($\uparrow$) & Acc.($\uparrow$) & Rob. ($\downarrow$) & EFO ($\uparrow$) \\
	\hline
	MDNet & \first{0.3783} & \first{0.6033} & \first{0.6936} & 0.97\\ 
	DeepSRDCF & \second{0.3181} & 0.5637 & \third{1.0457} & 0.26 \\ 
	EBT & \third{0.3130} & 0.4732 & \second{1.0213} & 2.74\\
	\textbf{MemDTC*} & 0.3005 & 0.5646 & 1.4610  & \rt{22.18}\\
	\textbf{MemDTC} & 0.2948 & 0.5509 & 1.6365 & \rt{21.95}\\
	SiamFC\footnotemark & 0.2889 & 0.5335 & - & - \\
	SRDCF & 0.2877 & 0.5592 & 1.2417 &1.36 \\
	\textbf{MemTrack*} & 0.2842 & 0.5573 &1.6768 &  \rt{\second{26.24}}\\
	LDP & 0.2785 & 0.4890 & 1.3332 &5.17\\
	sPST & 0.2767 & 0.5473 & 1.4796 & 1.16\\
	\textbf{MemTrack} & 0.2753 & 0.5582 & 1.7286 & \rt{\third{26.11}}\\
	SC-EBT & 0.2548 & 0.5529 & 1.8587 &1.83\\
	NSAMF & 0.2536 & 0.5305 & 1.2921 &6.81\\
	Struck & 0.2458 & 0.4712 & 1.6097 &3.52\\
	RAJSSC & 0.2420 & \third{0.5659} & 1.6296 &2.67\\
	S3Tracker & 0.2403 & 0.5153 & 1.7680 & \rt{20.04}\\
	SumShift & 0.2341 & 0.5169 & 1.6815 & \rt{23.55}\\
	SODLT & 0.2329 & 0.5607 & 1.7769 &1.14\\
	DAT & 0.2238 & 0.4856 & 2.2583 &14.87\\
	MEEM & 0.2212 & 0.4993 & 1.8535&3.66 \\
	RobStruck & 0.2198 & 0.4793 & 1.4724 &1.67\\
	OACF & 0.2190 & \second{0.5751} & 1.8128 &9.88\\
	MCT & 0.2188 & 0.4703 & 1.7609 &3.98\\
	HMMTxD & 0.2185 & 0.5278 & 2.4835 &2.17\\
	ASMS & 0.2117 & 0.5066 & 1.8464&\rt{\first{142.26}} \\
	MKCF+ & 0.2095 & 0.5153 & 1.8318 &1.79\\
	TRIC-track & 0.2088 & 0.4618 & 2.3426 &0.03\\
	AOG & 0.2080 & 0.5067 & 1.6727 &1.26\\
	SME & 0.2068 & 0.5528 & 1.9763 &5.77\\
	MvCFT & 0.2059 & 0.5220 & 1.7220 &11.85\\\hline
\end{tabular}
\egroup
\end{center}
\caption{Results on VOT-2015. The evaluation metrics include expected average overlap (EAO), accuracy value (Acc.), robustness value (Rob.) and equivalent filter operations (EFO). The top three performing trackers are colored with red, green and blue respectively. The up arrows indicate higher values are better for that metric, while down arrows mean lower values are better.
\abcn{Real-time methods ($>$15 EFO) are underlined.}
}
\label{tb:2}
\vspace{-8mm}
\end{table}
\footnotetext{Results are obtained from the original SiamFC paper \cite{Bertinetto2016}.}

\subsubsection{VOT-2016 Results} 
\begin{figure}[t]
	\begin{center}
		\includegraphics[width=0.49\linewidth]{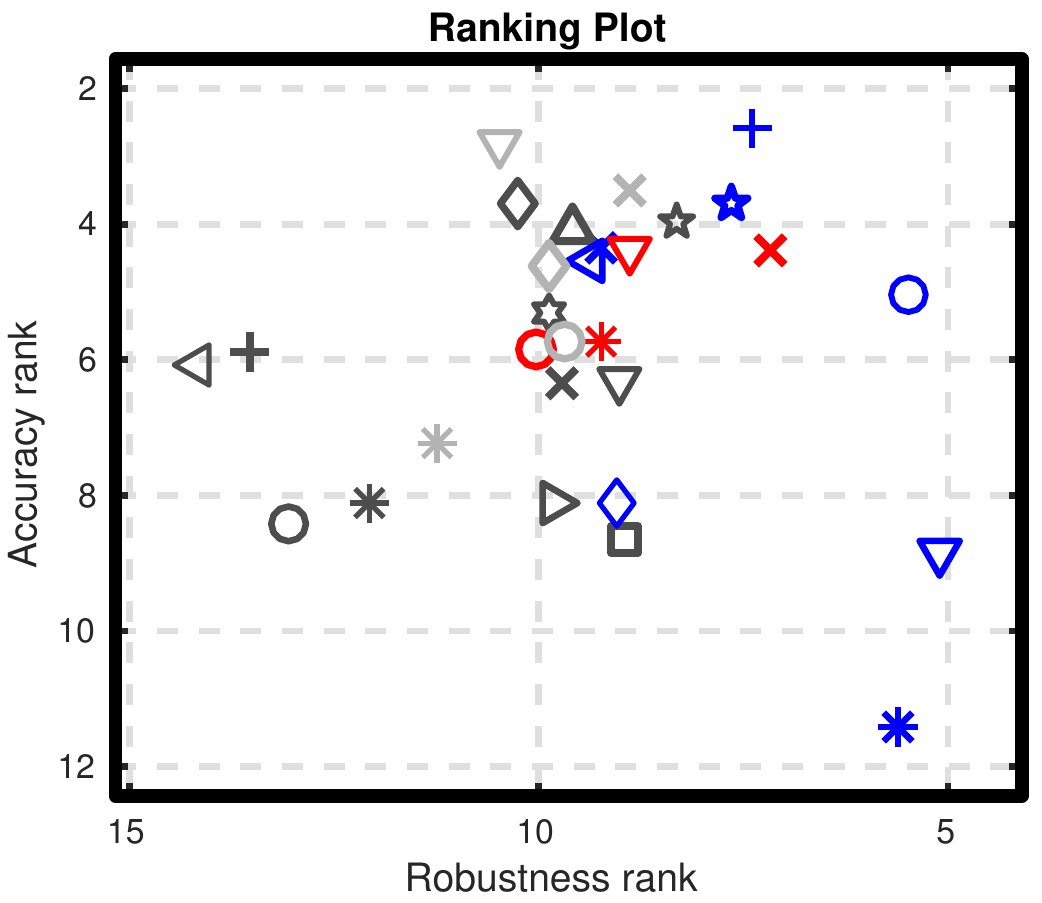}
		\includegraphics[width=0.49\linewidth]{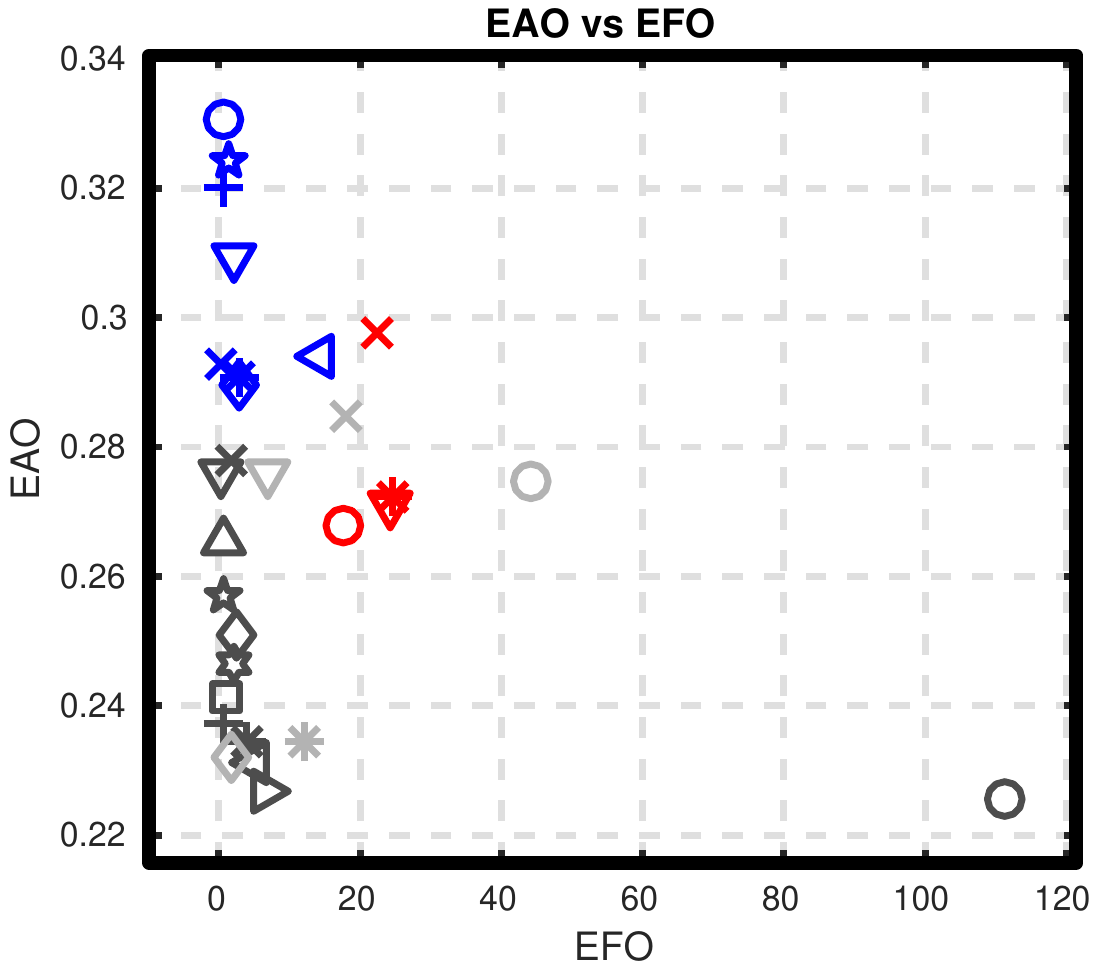}
		\includegraphics[width=0.9\linewidth]{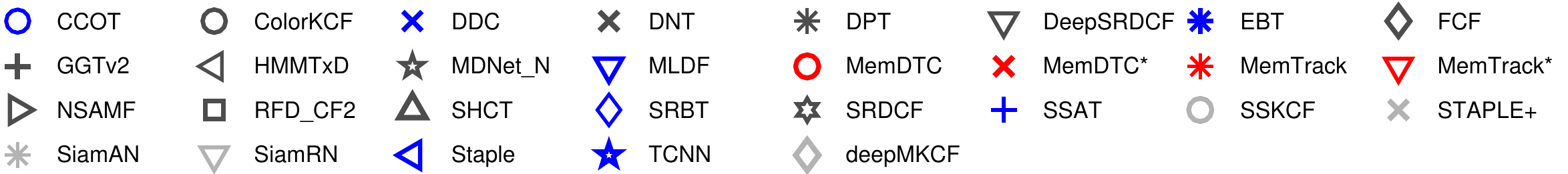}
	\end{center}
	\vspace{-3mm}
	\caption{The AR rank plots and EAO vs. EFO 
	for VOT-2016.  See the caption of Figure \ref{fig:13} for description of the colors.}
	\label{fig:14}
	\vspace{-4mm}
\end{figure}

Together 48 tracking methods are submitted to the VOT-2016 challenge and 22 baseline algorithms are provided by the VOT-2016 committee and associates. Table \ref{tb:3} summarizes the detailed comparison results with the top 25 performing trackers. Overall, CCOT \cite{Danelljan2016-1} achieves the best results on EAO,  SSAT \cite{Kristan2016} obtains the best performance on accuracy value, and TCNN \cite{Nam2016-1} outperforms all other trackers on robustness value. Our MemDTC* ranks the 5th over EAO, and surpasses other variants MemTrack, MemTrack* and MemDTC by a large margin. With the use of deeper networks, VGG, SSAT and MLDF achieve better EAO compared with MemDTC*, which is however at the cost of consuming considerable computation leading to non-realtime speed.  It is worth noting that the proposed MemDTC* runs much faster than those trackers that rank ahead of it. Figure \ref{fig:14} shows the accuracy-robustness ranking plot (left) and EAO vs.~EFO plot (right) on the VOT-2016 dataset. Our algorithm MemDTC* achieves better robustness rank than the other three variants MemDTC, MemTrack* and MemTrack.
As reported in VOT-2016, the SOTA bound is EAO 0.251, which all our trackers exceed. In addition, our trackers outperform the baseline matching-based tracker SiamAN (SiamFC with AlexNet), SiamRN (SiamFC with ResNet) and RFL \cite{Yang2017} over all evaluation metrics. 

\begin{table}
	\small
	\begin{center}
		\bgroup
		\def\arraystretch{1.15}
		\begin{tabular}{ccccc}
			\hline
			\multirow{2}{*}{\textbf{ }}
			& EAO ($\uparrow$) & Acc.($\uparrow$) & Rob. ($\downarrow$) & EFO ($\uparrow$) \\
			\hline
			CCOT & \first{0.3305} & 0.5364 & \second{0.8949} & 0.51\\
			TCNN & \second{0.3242} & \third{0.5530} & \first{0.8302} &1.35\\
			SSAT & \third{0.3201} & \first{0.5764} & 1.0462 &0.80\\
			MLDF & 0.3093 & 0.4879 & 0.9236 &2.20\\
			\textbf{MemDTC*} & 0.2976 & 0.5297 & 1.3106 &\rt{22.30}\\
			Staple & 0.2940 & 0.5406 & 1.4158 &14.43\\
			DDC & 0.2928 & 0.5363 & 1.2656 &0.16\\
			EBT & 0.2909 & 0.4616 & 1.0455 &2.87\\
			SRBT & 0.2897 & 0.4949 & 1.3314 &2.90\\
			STAPLE+ & 0.2849 & \second{0.5537} & 1.3094&\rt{18.12} \\
			DNT & 0.2781 & 0.5136 & 1.2004 &1.88\\
			SiamRN & 0.2760 & 0.5464 & 1.3617 &7.05\\
			DeepSRDCF & 0.2759 & 0.5229 & 1.2254 &0.38\\
			SSKCF & 0.2747 & 0.5445 & 1.4299 &\rt{\second{44.06}}\\
			\textbf{MemTrack} & 0.2723 & 0.5273 & 1.4381 &\rt{\third{24.60}}\\
			\textbf{MemTrack*} & 0.2713 & 0.5378 & 1.4736 & \rt{24.22}\\
			\textbf{MemDTC} & 0.2679 & 0.5109 & 1.8287 & \rt{21.42}\\
			SHCT & 0.2654 & 0.5431 & 1.3902 &0.54\\
			MDNet\_N & 0.2569 & 0.5396 & \third{0.9123} &0.69\\
			FCF & 0.2508 & 0.5486 & 1.8460 &2.39\\
			SRDCF & 0.2467 & 0.5309 & 1.4332 & 1.99\\
			RFD\_CF2 & 0.2414 & 0.4728 & 1.2697 &1.20\\
			GGTv2 & 0.2373 & 0.5150 & 1.7334 &0.52\\
			SiamAN & 0.2345 & 0.5260 & 1.9093 &11.93\\
			DPT & 0.2343 & 0.4895 & 1.8509 &4.03\\
			deepMKCF & 0.2320 & 0.5405 & 1.2271 &1.89\\
			HMMTxD & 0.2311 & 0.5131 & 2.1444 &4.99\\
			NSAMF & 0.2267 & 0.4984 & 1.2536 &6.61\\
			ColorKCF & 0.2257 & 0.5003 & 1.5009 &\rt{\first{111.39}}\\\hline
		\end{tabular}
		\egroup
	\end{center}
	\caption{Results on VOT-2016. 
	See the caption of Table \ref{tb:2} for more information.
}
\vspace{-5mm}
	\label{tb:3}
\end{table}

\subsubsection{VOT-2017 Results} 
\begin{figure}[t]
	\begin{center}
		\includegraphics[width=0.49\linewidth]{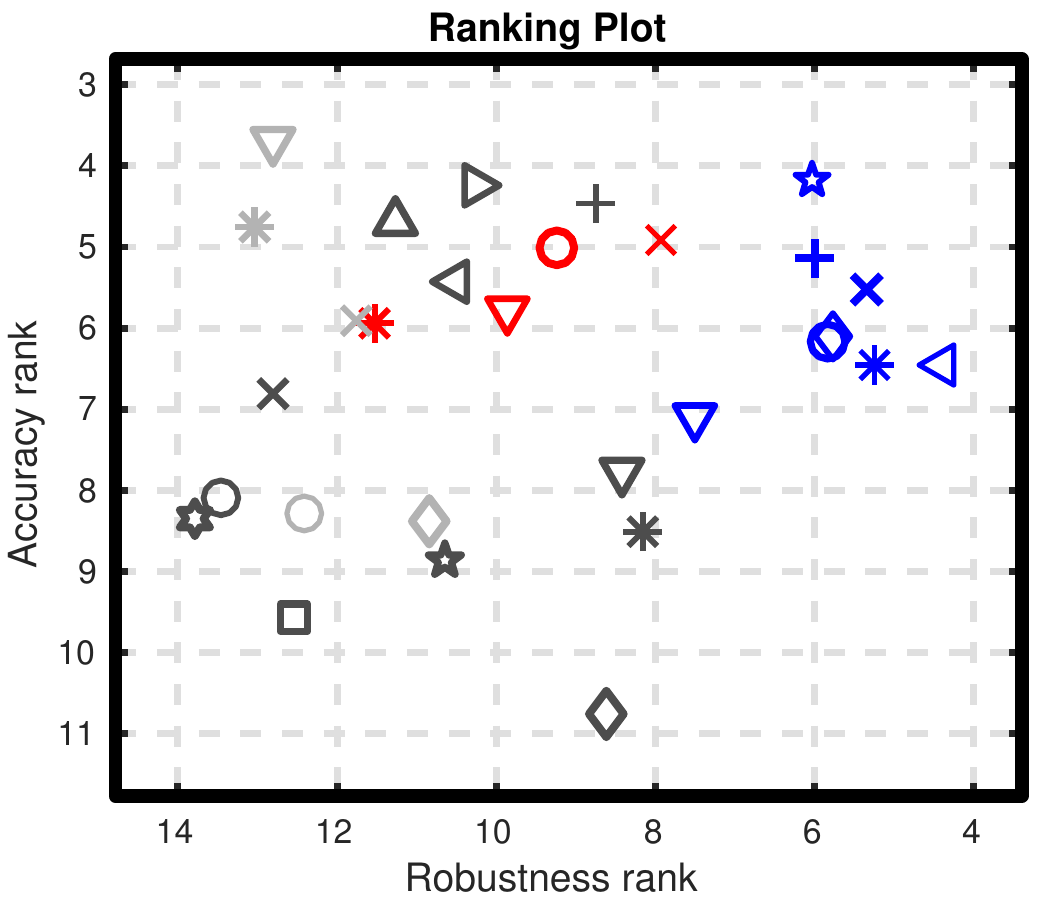}
		\includegraphics[width=0.49\linewidth]{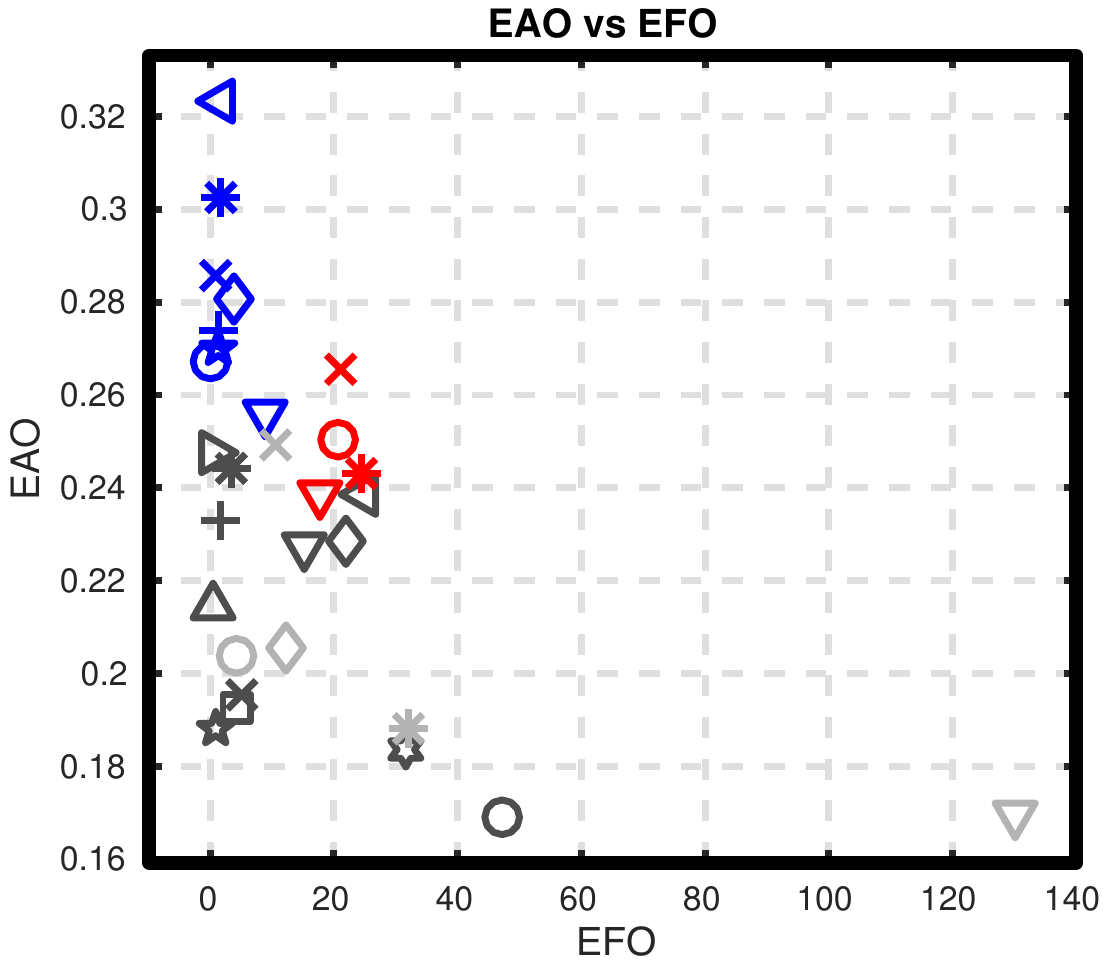}
		\includegraphics[width=0.9\linewidth]{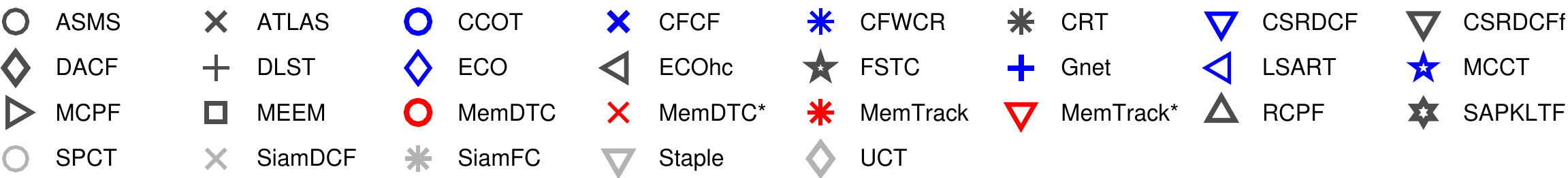}
	\end{center}
	\vspace{-3mm}
	\caption{The AR rank plots and \abc{EAO vs. EFO} 
	for VOT-2017. Our methods are colored with red while the top 10 methods are marked with blue. Others are colored with gray.}
	\label{fig:15}
	\vspace{-4mm}
\end{figure}

There are 38 valid entries submitted to the VOT-2017 challenge and 13 baseline trackers contributed by the VOT-2017 committee and associates. 
Table \ref{tb:4} shows the comparison results of the top 25 performing trackers, as well as our proposed methods.
  The winner of VOT-2017 is LSART \cite{Sun2018}, which utilizes a weighted cross-patch similarity kernel for kernelized ridged regression and a fully convolutional neural network with spatially regularized kernels. However due to heavy model fusing and the use of deeper networks (VGGNet), it runs at $\sim$2 fps. ECO \cite{Danelljan2017}, which ranks the fourth place on EAO, improves CCOT \cite{Danelljan2016-1} over both performance and speed by introducing a factorized convolution operator. However the speed of ECO is still far from real-time even though it is faster than CCOT. CFWCR \cite{Kristan2017} adopts ECO as the baseline and further boosts it by using more layers for feature fusing. CFCF \cite{Gundogdu2018} also utilizes ECO as the baseline tracker and improves it by selecting different layer features (first, fifth and sixth) of a newly trained fully convolutional network on ILSVRC \cite{ILSVRC15} as the input of ECO. In addition, Gnet \cite{Kristan2017} integrates GoogLeNet features into SRDCF \cite{Danelljan2015} and ECO \cite{Danelljan2017}.
  Since those trackers that rank ahead of our MemDTC* are usually based on either ECO or SRDCF using deeper networks (like VGG), which are thus not real-time, our tracker performs favorably against these top performing trackers, while retaining real-time speed. Furthermore, our MemDTC* outperforms both our earlier work MemTrack \cite{Yang2018} and the baseline tracker SiamFC \cite{Bertinetto2016}. Figure \ref{fig:15} shows the accuracy-robustness ranking plot (left) and EAO vs.~EFO plot (right) on the VOT-2017 dataset. Our methods perform favorably against state-of-the-art trackers in terms of both accuracy and robustness, \abcn{and has the best performance among real-time trackers.}

\begin{table}
	\small
	\begin{center}
		\bgroup
		\def\arraystretch{1.15}
		\begin{tabular}{ccccc}
			\hline
			\multirow{2}{*}{\textbf{ }}
			& EAO ($\uparrow$) & Acc.($\uparrow$) & Rob. ($\downarrow$) & EFO ($\uparrow$) \\
			\hline
			LSART & \first{0.3211} & 0.4913 & \first{0.9432} &1.72\\
			CFWCR & \second{0.2997} & 0.4818 & 1.2103 &1.80\\
			CFCF & \third{0.2845} & 0.5042 & 1.1686 &0.85\\
			ECO & 0.2803 & 0.4806 & \third{1.1167} &3.71\\
			Gnet & 0.2723 & 0.4992 & \second{0.9973} &1.29\\
			MCCT & 0.2679 & \second{0.5198} & 1.1258 &1.32\\
			CCOT & 0.2658 & 0.4887 & 1.3153 &0.15\\
			\textbf{MemDTC*} & 0.2651 & 0.4909 & 1.5287 & \rt{21.12}\\
			CSRDCF & 0.2541 & 0.4835 & 1.3095 &8.75\\
			\textbf{MemDTC} & 0.2504 & 0.4924 & 1.7730 & \rt{20.49}\\
			SiamDCF & 0.2487 & 0.4956 & 1.8659 &10.73\\
			MCPF & 0.2477 & 0.5081 & 1.5903 &0.42\\
			CRT & 0.2430 & 0.4613 & 1.2367 &3.24\\
			\textbf{MemTrack} & 0.2427 & 0.4935 & 1.7735 & \rt{24.27}\\
			\textbf{MemTrack*} & 0.2416 & 0.5025 & 1.8058 & \rt{24.74}\\
			ECOhc & 0.2376 & 0.4905 & 1.7737 & \rt{17.71}\\
			DLST & 0.2329 & 0.5051 & 1.5667 &1.89\\
			DACF & 0.2278 & 0.4498 & 1.3211 & \rt{21.96}\\
			CSRDCFf & 0.2257 & 0.4712 & 1.3905 & \rt{15.05}\\
			RCPF & 0.2144 & 0.5001 & 1.5892 &0.42\\
			UCT & 0.2049 & 0.4839 & 1.8307 &12.20\\
			SPCT & 0.2025 & 0.4682 & 2.1547 &4.40\\
			ATLAS & 0.1953 & 0.4821 & 2.5702 &5.21\\
			MEEM & 0.1914 & 0.4548 & 2.1111 &4.12\\
			FSTC & 0.1878 & 0.4730 & 1.9235 &0.96\\
			SiamFC & 0.1876 & 0.4945 & 2.0485 &\rt{\third{31.89}}\\
			SAPKLTF & 0.1835 & 0.4764 & 2.2002 & \rt{31.65}\\
			ASMS & 0.1687 & 0.4868 & 2.2496 & \rt{\first{130.02}}\\
			Staple & 0.1685 & \third{0.5194} & 2.5068 &\rt{\second{47.01}}\\\hline
		\end{tabular}
		\egroup
	\end{center}
	\caption{Results on VOT-2017. 
	See the caption of Table \ref{tb:2} for more information.
	}
	\label{tb:4}
	\vspace{-5mm}
\end{table}

\subsection{Ablation Studies}\label{abla}

\begin{figure}[t]
	\begin{center}
		\includegraphics[width=\linewidth]{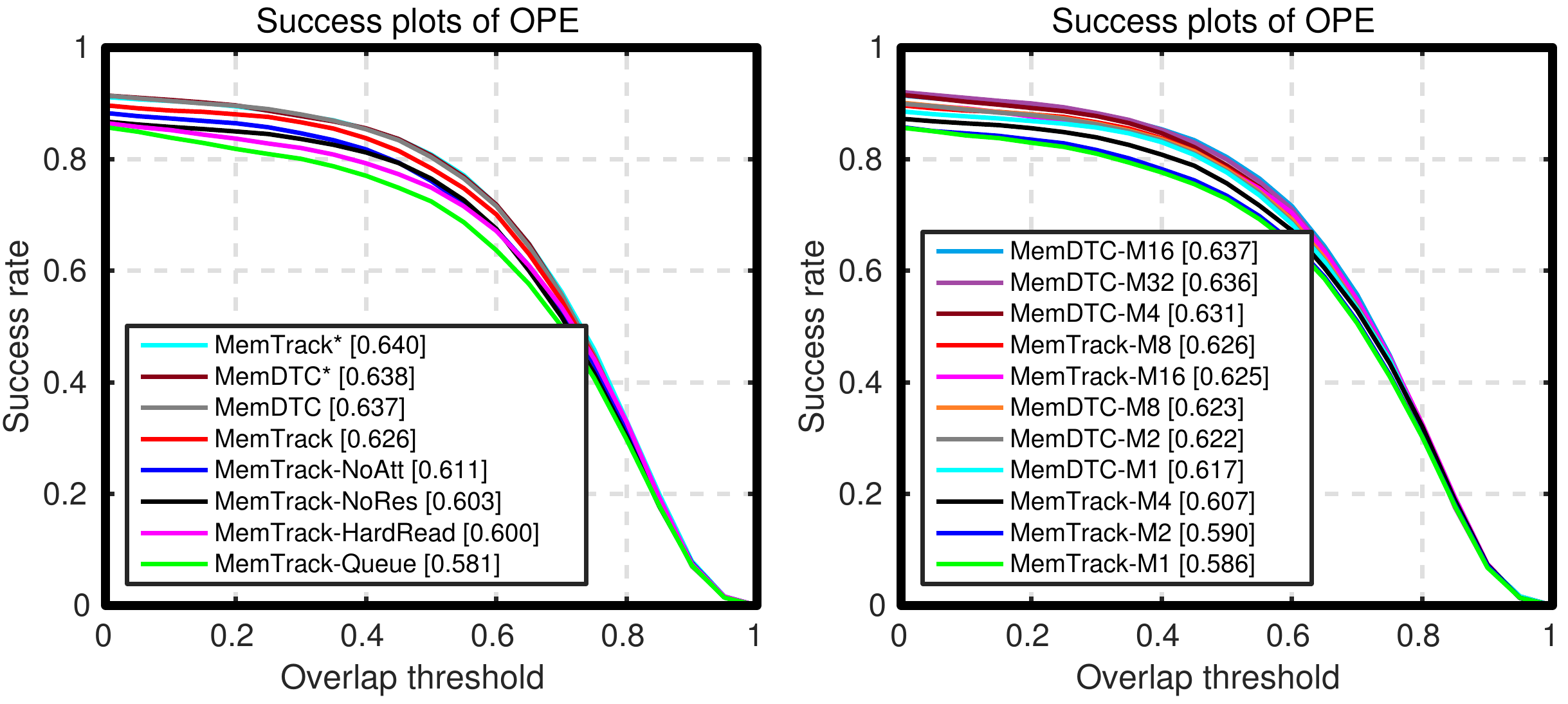}
	\end{center}
	\vspace{-4mm}
	\caption{Ablation studies: (left) success plots of different variants of our tracker on OTB-2015; (right) success plots for different memory sizes \{1, 2, 4, 8, 16\} for positive memory slot, \yty{\{1, 2, 4, 8, 16, 32\} for negative memory slot} on OTB-2015. }
	\vspace{-4mm}
	\label{fig:7}
\end{figure}

Our preliminary tracker MemTrack \cite{Yang2018} contains three important components: 1) an attention mechanism, which calculates the attended feature vector for memory reading; 2) a dynamic memory network, which maintains the target's appearance variations; and 3) residual template learning, which controls the amount of model updating for each channel of the template. To evaluate their separate contributions to our tracker, we implement several variants of our method and verify them on OTB-2015 dataset.
The results of the ablation study are presented in Figure \ref{fig:7} (left).

We first design a variant of MemTrack without attention mechanism (MemTrack-NoAtt), which averages all $L$ feature vectors to get the feature vector $\mathbf{a}_t$ for the LSTM input. 
Mathematically, it changes 
(2) to $\mathbf{a}_t = \frac{1}{L}\sum_{i=1}^{L}\mathbf{f}^*_{t,i} $. 
Memtrack-NoAtt 
decreases performance (see Figure \ref{fig:7} left), which shows the benefit of using attention to roughly localize the target in the search image.
We also design a naive strategy that simply writes the new target template sequentially into the memory slots as a queue (MemTrack-Queue). When the memory is fully occupied, the oldest template will be replaced with the new  template. The retrieved template is generated by averaging all templates stored in the memory slots. 
This simple approach cannot produce good performance (Figure \ref{fig:7} left), which shows the necessity of our dynamic memory network. \yty{We devise a hard template reading scheme (MemTrack-HardRead), \emph{i.e.}, retrieving a single template by max cosine distance, to replace the soft weighted sum reading scheme. 
This design decreases the performance (Figure \ref{fig:7} left), most likely because the non-differentiable reading leads to an inferior model.}
To verify the effectiveness of gated residual template learning, we design another variant of MemTrack--- removing channel-wise residual gates (MemTrack-NoRes), \emph{i.e.} directly adding the retrieved and initial templates to get the final template.
Our gated residual template learning mechanism boosts the performance (Figure \ref{fig:7} left) as it helps to select correct residual channel features for template updating.

\yty{In this paper, we improve MemTrack with two techniques: distractor template canceling and auxiliary classification loss. Our newly proposed methods MemTrack*, MemDTC and MemDTC* consistently outperform our earlier work MemTrack (Figure~\ref{fig:7} left).} \ytyy{Without the auxiliary classification loss, MemDTC outperforms MemTrack on both the precision and success plots of OTB-2013/2015, which demonstrates its effectiveness of the distractor template canceling strategy. When using the auxiliary classification loss, MemDTC* has slightly better performance that MemTrack* on OTB-2013, but slightly worse performance on OTB-2015. It is possible that the discrimination ability of the feature extractor (only a 5-layer CNN) limits the performance gain of the auxiliary loss. We also note that MemTrack* achieves slightly worse EAO than MemTrack on VOT-2016/2017, while MemDTC* is better than MemDTC on VOT-2015/2016/2017. On these datasets, the auxiliary task cannot further improve the performance until the distractor template canceling scheme is used.}

We also investigate the effect of memory size on tracking performance. Figure \ref{fig:7} (right) shows the success plot on OTB-2015 using different numbers of memory slots. For  MemTrack, tracking accuracy increases along with the memory size and saturates at 8 memory slots. Considering the runtime and memory usage, we choose 8 as the default number of positive memory slots. \yty{For our improved tracker MemDTC, we keep the number of positive memory slot fixed at 8, and change the number of negative memory slots. The tracking performance increases with the number of negative memory slot, \abc{and saturates at} 
16.}

\section{Conclusion}
In this paper, we propose a dynamic memory network with an external addressable memory block for visual tracking, aiming to adapt matching templates to object appearance variations. 
An LSTM with attention scheme controls the memory access by parameterizing the memory interactions. We develop channel-wise gated residual template learning to form the \yty{positive} matching model, which preserves the conservative information present in the initial target, while providing online adapability of each feature channel. \yty{To alleviate the drift problem caused by distractor targets, we devise a distractor template canceling scheme \abc{that inhibits channels in the final template that are not discriminative.} Furthermore, we improve the tracking performance by introducing an auxiliary classification loss branch after the feature extractor,  aiming to learn semantic features that complement the features trained by similarity matching.} Once the offline training process is finished, no online fine-tuning is needed, which leads to real-time speed. Extensive experiments on standard tracking benchmark demonstrates the effectiveness of our proposed trackers.

\section*{Acknowledgments}

This work was supported by grants from the Research Grants Council of the Hong Kong Special Administrative Region, China 
(CityU 11200314, and CityU 11212518). We grateful for the support of NVIDIA Corporation with the donation of the Tesla K40 GPU used for this research.

\bibliographystyle{IEEEtran}
\bibliography{egbib}

\begin{thebibliography}{10}
\providecommand{\url}[1]{#1}
\csname url@samestyle\endcsname
\providecommand{\newblock}{\relax}
\providecommand{\bibinfo}[2]{#2}
\providecommand{\BIBentrySTDinterwordspacing}{\spaceskip=0pt\relax}
\providecommand{\BIBentryALTinterwordstretchfactor}{4}
\providecommand{\BIBentryALTinterwordspacing}{\spaceskip=\fontdimen2\font plus
\BIBentryALTinterwordstretchfactor\fontdimen3\font minus
  \fontdimen4\font\relax}
\providecommand{\BIBforeignlanguage}[2]{{%
\expandafter\ifx\csname l@#1\endcsname\relax
\typeout{** WARNING: IEEEtran.bst: No hyphenation pattern has been}%
\typeout{** loaded for the language `#1'. Using the pattern for}%
\typeout{** the default language instead.}%
\else
\language=\csname l@#1\endcsname
\fi
#2}}
\providecommand{\BIBdecl}{\relax}
\BIBdecl

\bibitem{Krizhevsky2012}
A.~Krizhevsky, I.~Sutskever, and G.~E. Hinton, ``{Imagenet classification with
  deep convolutional neural networks},'' in \emph{NIPS}, 2012.

\bibitem{Szegedy2015}
C.~Szegedy, W.~Liu, Y.~Jia, and P.~Sermanet, ``{Going deeper with
  convolutions},'' in \emph{CVPR}, 2015.

\bibitem{He2016}
K.~He, X.~Zhang, S.~Ren, and J.~Sun, ``{Deep Residual Learning for Image
  Recognition},'' in \emph{CVPR}, 2016.

\bibitem{Girshick2014}
R.~Girshick, J.~Donahue, T.~Darrell, and J.~Malik, ``{Rich feature hierarchies
  for accurate object detection and semantic segmentation},'' in \emph{CVPR},
  2014.

\bibitem{Girshick2015}
R.~Girshick, ``{Fast R-CNN},'' in \emph{ICCV}, 2015.

\bibitem{Ren2015}
S.~Ren, K.~He, R.~Girshick, and J.~Sun, ``{Faster R-CNN: Towards Real-Time
  Object Detection with Region Proposal Networks},'' in \emph{NIPS}, 2015.

\bibitem{Long2015}
J.~Long, E.~Shelhamer, and T.~Darrell, ``{Fully convolutional networks for
  semantic segmentation},'' in \emph{CVPR}, 2015.

\bibitem{Noh2016}
H.~Noh, S.~Hong, and B.~Han, ``{Learning deconvolution network for semantic
  segmentation},'' in \emph{ICCV}, 2016.

\bibitem{Li2017}
Y.~Li, H.~Qi, J.~Dai, X.~Ji, and Y.~Wei, ``{Fully Convolutional Instance-aware
  Semantic Segmentation},'' in \emph{CVPR}, 2017.

\bibitem{Song2017}
Y.~Song, C.~Ma, L.~Gong, J.~Zhang, R.~Lau, and M.-H. Yang, ``{CREST:
  Convolutional Residual Learning for Visual Tracking},'' in \emph{ICCV}, 2017.

\bibitem{Nam2016}
H.~Nam and B.~Han, ``{Learning Multi-Domain Convolutional Neural Networks for
  Visual Tracking},'' in \emph{CVPR}, 2016.

\bibitem{Wang2015}
L.~Wang, W.~Ouyang, X.~Wang, and H.~Lu, ``{Visual Tracking with Fully
  Convolutional Networks},'' in \emph{ICCV}, 2015.

\bibitem{Bertinetto2016}
L.~Bertinetto, J.~Valmadre, J.~F. Henriques, A.~Vedaldi, and P.~H.~S. Torr,
  ``{Fully-Convolutional Siamese Networks for Object Tracking},'' in
  \emph{ECCVW}, 2016.

\bibitem{Guo2017}
Q.~Guo, W.~Feng, C.~Zhou, R.~Huang, L.~Wan, and S.~Wang, ``{Learning Dynamic
  Siamese Network for Visual Object Tracking},'' in \emph{ICCV}, 2017.

\bibitem{Tao2016}
R.~Tao, E.~Gavves, and A.~W.~M. Smeulders, ``{Siamese Instance Search for
  Tracking},'' in \emph{CVPR}, 2016.

\bibitem{Held2016}
D.~Held, S.~Thrun, and S.~Savarese, ``{Learning to Track at 100 FPS with Deep
  Regression Networks},'' in \emph{ECCV}, 2016.

\bibitem{Valmadre2017}
J.~Valmadre, L.~Bertinetto, F.~Henriques, A.~Vedaldi, and P.~H.~S. Torr,
  ``{End-to-end representation learning for Correlation Filter based
  tracking},'' in \emph{CVPR}, 2017.

\bibitem{Nam2016-1}
H.~Nam, M.~Baek, and B.~Han, ``{Modeling and Propagating CNNs in a Tree
  Structure for Visual Tracking},'' in \emph{arXiv}, 2016.

\bibitem{Yang2017}
T.~Yang and A.~B. Chan, ``{Recurrent Filter Learning for Visual Tracking},'' in
  \emph{ICCVW}, 2017.

\bibitem{He2018}
A.~He, C.~Luo, X.~Tian, and W.~Zeng, ``{A Twofold Siamese Network for Real-Time
  Object Tracking},'' in \emph{CVPR}, 2018.

\bibitem{Wang2018}
Q.~Wang, Z.~Teng, J.~Xing, J.~Gao, W.~Hu, and S.~Maybank, ``{Learning
  Attentions : Residual Attentional Siamese Network for High Performance Online
  Visual Tracking},'' in \emph{CVPR}, 2018.

\bibitem{Yang2018}
T.~Yang and A.~B. Chan, ``{Learning Dynamic Memory Networks for Object
  Tracking},'' in \emph{ECCV}, 2018.

\bibitem{Grabner2008}
H.~Grabner, C.~Leistner, and H.~Bischof, ``{Semi-Supervised On-line Boosting
  for Robust Tracking},'' in \emph{ECCV}, 2008.

\bibitem{Babenko2011}
B.~Babenko, S.~Member, M.-h. Yang, and S.~Member, ``{Robust Object Tracking
  with Online Multiple Instance Learning},'' \emph{TPAMI}, 2011.

\bibitem{Kalal2012}
Z.~Kalal, K.~Mikolajczyk, and J.~Matas, ``{Tracking-Learning-Detection},''
  \emph{TPAMI}, 2012.

\bibitem{li2018deep}
P.~Li, D.~Wang, L.~Wang, and H.~Lu, ``Deep visual tracking: Review and
  experimental comparison,'' \emph{Pattern Recognition}, 2018.

\bibitem{Srivastava2014}
N.~Srivastava, G.~E. Hinton, A.~Krizhevsky, I.~Sutskever, and R.~Salakhutdinov,
  ``{Dropout : A Simple Way to Prevent Neural Networks from Overfitting},''
  \emph{JMLR}, 2014.

\bibitem{Han2017}
B.~Han, J.~Sim, and H.~Adam, ``{BranchOut: Regularization for Online Ensemble
  Tracking with Convolutional Neural Networks},'' in \emph{CVPR}, 2017.

\bibitem{Huang2017}
C.~Huang, S.~Lucey, and D.~Ramanan, ``{Learning Policies for Adaptive Tracking
  with Deep Feature Cascades},'' in \emph{ICCV}, 2017.

\bibitem{chi2017dual}
Z.~Chi, H.~Li, H.~Lu, and M.-H. Yang, ``Dual deep network for visual
  tracking,'' \emph{TIP}, 2017.

\bibitem{Graves2014}
A.~Graves, G.~Wayne, and I.~Danihelka, ``{Neural Turing Machines},''
  \emph{Arxiv}, 2014.

\bibitem{Weston2015}
J.~Weston, S.~Chopra, and A.~Bordes, ``{Memory Networks},'' in \emph{ICLR},
  2015.

\bibitem{Sukhbaatar2015}
S.~Sukhbaatar, A.~Szlam, J.~Weston, and R.~Fergus, ``{End-To-End Memory
  Networks},'' in \emph{NIPS}, 2015.

\bibitem{Graves2016}
A.~Graves, G.~Wayne, M.~Reynolds, T.~Harley, I.~Danihelka,
  A.~Grabska-Barwi{\'{n}}ska, S.~{G{\'{o}}mez Colmenarejo}, E.~Grefenstette,
  T.~Ramalho, J.~Agapiou, A.~P. Badia, K.~{Moritz Hermann}, Y.~Zwols,
  G.~Ostrovski, A.~Cain, H.~King, C.~Summerfield, P.~Blunsom, K.~Kavukcuoglu,
  and D.~Hassabis, ``{Hybrid computing using a neural network with dynamic
  external memory},'' \emph{Nature}, 2016.

\bibitem{Santoro2016}
A.~Santoro, S.~Bartunov, M.~Botvinick, D.~Wierstra, and T.~Lillicrap,
  ``{One-shot Learning with Memory-Augmented Neural Networks},'' in
  \emph{ICML}, 2016.

\bibitem{Liu2017}
B.~Liu, Y.~Wang, Y.-W. Tai, and C.-K. Tang, ``{MAVOT: Memory-Augmented Video
  Object Tracking},'' \emph{arXiv}, 2017.

\bibitem{collobert2008unified}
R.~Collobert and J.~Weston, ``{A Unified Architecture for Natural Language
  Processing: Deep Neural Networks with Multitask Learning},'' in \emph{ICML},
  2008.

\bibitem{deng2013new}
L.~Deng, G.~Hinton, and B.~Kingsbury, ``{New Types of Deep Neural Network
  Learning for Speech Recognition and Related Applications: An Overview},'' in
  \emph{ICASSP}, 2013.

\bibitem{girshick2015fast}
R.~Girshick, ``{Fast R-CNN},'' in \emph{ICCV}, 2015.

\bibitem{caruana1997multitask}
R.~Caruana, ``{Multitask Learning},'' \emph{Machine learning}, 1997.

\bibitem{zhang2012convex}
Y.~Zhang and D.-Y. Yeung, ``{A Convex Formulation for Learning Task
  Relationships in Multi-task Learning},'' \emph{arXiv}, 2012.

\bibitem{li2015heterogeneous}
S.~Li, Z.-Q. Liu, and A.~B. Chan, ``{Heterogeneous Multi-task Learning for
  Human Pose Estimation with Deep Convolutional Neural Network},'' \emph{IJCV},
  2015.

\bibitem{yao2012describing}
J.~Yao, S.~Fidler, and R.~Urtasun, ``{Describing the Scene as A Whole: Joint
  Object Detection, Scene Classification and Semantic Segmentation},'' in
  \emph{CVPR}, 2012.

\bibitem{He2017}
K.~He, G.~Gkioxari, P.~Dollar, and R.~Girshick, ``{Mask R-CNN},'' in
  \emph{ICCV}, 2017.

\bibitem{eigen2015predicting}
D.~Eigen and R.~Fergus, ``{Predicting Depth, Surface Normals and Semantic
  Labels with a Common Multi-scale Convolutional Architecture},'' in
  \emph{ICCV}, 2015.

\bibitem{kendall2018multi}
A.~Kendall, Y.~Gal, and R.~Cipolla, ``{Multi-task Learning using Uncertainty to
  Weigh Losses for Scene Geometry and Semantics},'' in \emph{CVPR}, 2018.

\bibitem{Ba2016}
J.~L. Ba, J.~R. Kiros, and G.~E. Hinton, ``{Layer Normalization},''
  \emph{arXiv}, 2016.

\bibitem{Zeiler2014}
M.~D. Zeiler and R.~Fergus, ``{Visualizing and Understanding Convolutional
  Networks},'' in \emph{ECCV}, 2014.

\bibitem{Ma2015}
C.~Ma, J.-b. Huang, X.~Yang, and M.-h. Yang, ``{Hierarchical Convolutional
  Features for Visual Tracking},'' in \emph{ICCV}, 2015.

\bibitem{ILSVRC15}
O.~Russakovsky, J.~Deng, H.~Su, J.~Krause, S.~Satheesh, S.~Ma, Z.~Huang,
  A.~Karpathy, A.~Khosla, M.~Bernstein, A.~C. Berg, and L.~Fei-Fei, ``{ImageNet
  Large Scale Visual Recognition Challenge},'' \emph{IJCV}, 2015.

\bibitem{kingma2014adam}
D.~Kingma and J.~Ba, ``Adam: A method for stochastic optimization,''
  \emph{arXiv}, 2014.

\bibitem{abadi2016tensorflow}
M.~Abadi, A.~Agarwal, P.~Barham, E.~Brevdo, Z.~Chen, C.~Citro, G.~S. Corrado,
  A.~Davis, J.~Dean, M.~Devin \emph{et~al.}, ``Tensorflow: Large-scale machine
  learning on heterogeneous distributed systems,'' \emph{arXiv}, 2016.

\bibitem{Wu2013}
Y.~Wu, J.~Lim, and M.-H. Yang, ``{Online Object Tracking: A Benchmark},'' in
  \emph{CVPR}, 2013.

\bibitem{Wu2015}
------, ``{Object Tracking Benchmark},'' \emph{PAMI}, 2015.

\bibitem{Kristan2015}
M.~Kristan, R.~Pflugfelder, A.~Leonardis, J.~Matas, L.~{\v{C}}ehovin,
  G.~Nebehay, T.~Voj{\'{i}}$\backslash$vr, G.~Fernandez, and Others, ``{The
  visual object tracking VOT2015 challenge results},'' in \emph{ICCVW}, 2015.

\bibitem{Kristan2016}
M.~Kristan, A.~Leonardis, J.~Matas, and M.~Felsberg, ``{The Visual Object
  Tracking VOT2016 Challenge Results},'' in \emph{ECCVW}, 2016.

\bibitem{Kristan2017}
M.~Kristan, A.~Leonardis, J.~Matas, M.~Felsberg, R.~Pflugfelder, L.~{\v{C}}.
  Zajc, T.~Voj{\'{i}}r, G.~H{\"{a}}ger, A.~Luke{\v{z}}i{\v{c}}, A.~Eldesokey,
  G.~Fern{\'{a}}ndez, and Others, ``{The Visual Object Tracking VOT2017
  Challenge Results},'' in \emph{ICCVW}, 2017.

\bibitem{Li2018}
B.~Li, J.~Yan, W.~Wu, Z.~Zhu, and X.~Hu, ``{High Performance Visual Tracking
  with Siamese Region Proposal Network},'' in \emph{CVPR}, 2018.

\bibitem{Fan2017}
H.~Fan and H.~Ling, ``{Parallel Tracking and Verifying : A Framework for
  Real-Time and High Accuracy Visual Tracking},'' in \emph{ICCV}, 2017.

\bibitem{Wang2017}
M.~Wang, Y.~Liu, and Z.~Huang, ``{Large Margin Object Tracking with Circulant
  Feature Maps},'' in \emph{CVPR}, 2017.

\bibitem{Choi2017}
J.~Choi, H.~J. Chang, S.~Yun, T.~Fischer, Y.~Demiris, and {Jin Young Choi},
  ``{Attentional Correlation Filter Network for Adaptive Visual Tracking},'' in
  \emph{CVPR}, 2017.

\bibitem{Bertinetto2016-1}
L.~Bertinetto, J.~Valmadre, S.~Golodetz, O.~Miksik, and P.~Torr, ``{Staple:
  Complementary Learners for Real-Time Tracking},'' in \emph{CVPR}, 2016.

\bibitem{Danelljan2014}
M.~Danelljan, G.~H{\"{a}}ger, F.~Khan, and M.~Felsberg, ``{Accurate Scale
  Estimation for Robust Visual Tracking},'' in \emph{BMVC}, 2014.

\bibitem{Henriques2015}
J.~F. Henriques, R.~Caseiro, P.~Martins, and J.~Batista, ``{High-speed tracking
  with kernelized correlation filters},'' \emph{TPAMI}, 2015.

\bibitem{Lukezic2017}
A.~Luke{\v{z}}i{\v{c}}, T.~Voj{\'{i}}ř, L.~{\v{C}}ehovin, J.~Matas, and
  M.~Kristan, ``{Discriminative Correlation Filter with Channel and Spatial
  Reliability},'' in \emph{CVPR}, 2017.

\bibitem{Zhang2017}
T.~Zhang, C.~Xu, and M.-h. Yang, ``{Multi-task Correlation Particle Filter for
  Robust Object Tracking},'' in \emph{CVPR}, 2017.

\bibitem{Danelljan2016}
M.~Danelljan, G.~H{\"{a}}ger, F.~S. Khan, and M.~Felsberg, ``{Adaptive
  Decontamination of the Training Set: A Unified Formulation for Discriminative
  Visual Tracking},'' in \emph{CVPR}, 2016.

\bibitem{Danelljan2015}
M.~Danelljan, H.~Gustav, F.~S. Khan, and M.~Felsberg, ``{Learning Spatially
  Regularized Correlation Filters for Visual Tracking},'' in \emph{ICCV}, 2015.

\bibitem{Qi2016}
Y.~Qi, S.~Zhang, L.~Qin, H.~Yao, Q.~Huang, and J.~L. M.-h. Yang, ``{Hedged Deep
  Tracking},'' in \emph{CVPR}, 2016.

\bibitem{Danelljan2016-2}
M.~Danelljan, G.~Hager, F.~S. Khan, and M.~Felsberg, ``{Convolutional Features
  for Correlation Filter Based Visual Tracking},'' in \emph{ICCVW}, 2016.

\bibitem{Zhu2016}
G.~Zhu, F.~Porikli, and H.~Li, ``{Beyond Local Search: Tracking Objects
  Everywhere with Instance-Specific Proposals},'' in \emph{CVPR}, 2016.

\bibitem{Hua2015}
Y.~Hua, K.~Alahari, and C.~Schmid, ``{Online Object Tracking with Proposal
  Selection},'' in \emph{ICCV}, 2015.

\bibitem{Danelljan2016-1}
M.~Danelljan, A.~Robinson, F.~S. Khan, and M.~Felsberg, ``{Beyond correlation
  filters: Learning continuous convolution operators for visual tracking},'' in
  \emph{ECCV}, 2016.

\bibitem{Sun2018}
C.~Sun, H.~Lu, and M.-H. Yang, ``{Learning Spatial-Aware Regressions for Visual
  Tracking},'' in \emph{CVPR}, 2018.

\bibitem{Danelljan2017}
M.~Danelljan, G.~Bhat, F.~S. Khan, and M.~Felsberg, ``{ECO: Efficient
  Convolution Operators for Tracking},'' in \emph{CVPR}, 2017.

\bibitem{Gundogdu2018}
E.~Gundogdu and A.~A. Alatan, ``{Good Features to Correlate for Visual
  Tracking},'' \emph{TIP}, 2018.

\end{thebibliography}

%

\begin{IEEEbiography}[{\includegraphics[width=1in,height=1.25in,clip,keepaspectratio]{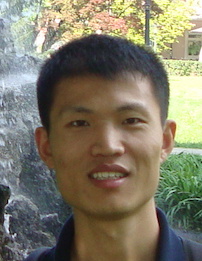}}]{Tianyu Yang}
	received the B.S. degree from
	Liaocheng University, China, and the
	M.Eng. degree from University of Chinese Academy of Sciences, China, in 2010 and 2013, respectively.
	He is currently a PhD student at City
	University of Hong Kong, China. His current research
	interests include visual tracking and deep
	learning.
\end{IEEEbiography}

\begin{IEEEbiography}[{\includegraphics[width=1in,height=1.25in,clip,keepaspectratio]{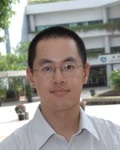}}]{Antoni B. Chan}
	received the B.S. and M.Eng.
	degrees in electrical engineering from Cornell
	University, Ithaca, NY, in 2000 and 2001, and
	the Ph.D. degree in electrical and computer engineering
	from the University of California, San
	Diego (UCSD), San Diego, in 2008. He is currently
	an Associate Professor in the Department
	of Computer Science, City University of Hong
	Kong. His research interests include computer
	vision, machine learning, pattern recognition,
	and music analysis.
\end{IEEEbiography}

\end{document}